\documentclass{article}

\usepackage{microtype}
\usepackage{graphicx}
\usepackage{subfigure}
\usepackage{booktabs} 
\usepackage{amsmath}
\usepackage{amssymb}
\usepackage{booktabs}
\usepackage{multirow}
\usepackage[dvipsnames]{xcolor}

\usepackage[hyphens]{url}
\usepackage{hyperref}
\hypersetup{breaklinks=true}

\newcommand{\eg}{{\it{e.g.}~}}
\newcommand{\ie}{{\it{i.e.}~}}

\usepackage[accepted]{icml2021}

\icmltitlerunning{Florence: A New Foundation Model for Computer Vision}

\usepackage[british,UKenglish,USenglish,english,american]{babel}

\usepackage[capitalize]{cleveref}
\crefname{section}{Sec.}{Secs.}
\Crefname{section}{Section}{Sections}
\Crefname{table}{Table}{Tables}
\crefname{table}{Tab.}{Tabs.}

\renewcommand{\comment}[1]{}

\newcommand{\Florence}{\emph{Florence~}}

\begin{document}

\twocolumn[
\icmltitle{Florence: A New Foundation Model for Computer Vision}



\icmlsetsymbol{team}{*}

\begin{icmlauthorlist}
\icmlauthor{Lu Yuan}{cogs}
\icmlauthor{Dongdong Chen}{team,cogs}
\icmlauthor{Yi-Ling Chen}{team,cogs}
\icmlauthor{Noel Codella}{team,cogs}
\icmlauthor{Xiyang Dai}{team,cogs}
\icmlauthor{Jianfeng Gao}{team,msr}
\icmlauthor{Houdong Hu}{team,cogs}
\icmlauthor{Xuedong Huang}{team,cogs}
\icmlauthor{Boxin Li}{team,cogs}
\icmlauthor{Chunyuan Li}{team,msr}
\icmlauthor{Ce Liu}{team,cogs}
\icmlauthor{Mengchen Liu}{team,cogs}
\icmlauthor{Zicheng Liu}{team,cogs}
\icmlauthor{Yumao Lu}{team,cogs}
\icmlauthor{Yu Shi}{team,cogs}
\icmlauthor{Lijuan Wang}{team,cogs}
\icmlauthor{Jianfeng Wang}{team,cogs}
\icmlauthor{Bin Xiao}{team,cogs}
\icmlauthor{Zhen Xiao}{team,cogs}
\icmlauthor{Jianwei Yang}{team,msr}
\icmlauthor{Michael Zeng}{team,cogs}
\icmlauthor{Luowei Zhou}{team,cogs}
\icmlauthor{Pengchuan Zhang}{team,msr}
\end{icmlauthorlist}

\icmlaffiliation{cogs}{Microsoft Cloud and AI}
\icmlaffiliation{msr}{Microsoft Research Redmond}

\icmlcorrespondingauthor{Lu Yuan}{luyuan@microsoft.com}


\vskip 0.3in
]

\printAffiliationsAndNotice{\icmlEqualContribution} 

\begin{abstract}

Automated visual understanding of our diverse and open world demands computer vision models to generalize well with minimal customization for specific tasks, similar to human vision. Computer vision foundation models, which are trained on diverse, large-scale dataset and can be adapted to a wide range of downstream tasks, are critical for this mission to solve real-world computer vision applications. While existing vision foundation models such as CLIP~\cite{radford2021learning}, ALIGN~\cite{jia2021scaling}, and Wu Dao 2.0~\cite{Wudao2} focus mainly on mapping images and textual representations to a cross-modal shared representation, we introduce a new computer vision foundation model, \emph{Florence}, to expand the representations from coarse (scene) to fine (object), from static (images) to dynamic (videos), and from RGB to multiple modalities (caption, depth). By incorporating universal visual-language representations from Web-scale image-text data, our \Florence model can be easily adapted for various computer vision tasks, such as classification, retrieval, object detection, VQA, image caption, video retrieval and action recognition. Moreover, \Florence demonstrates outstanding performance in many types of transfer learning: fully sampled fine-tuning, linear probing, few-shot transfer and zero-shot transfer for novel images and objects. All of these properties are critical for our vision foundation model to serve general purpose vision tasks. \Florence achieves new state-of-the-art results in majority of $44$ representative benchmarks, \eg ImageNet-1K zero-shot classification with top-1 accuracy of
{${83.74}$} and the top-5 accuracy of {${97.18}$}, ${62.4}$ mAP on COCO fine tuning, ${80.36}$ on VQA, and ${87.8}$ on Kinetics-600.
\end{abstract}

\section{Introduction}
\label{sec:intro}

Human-like AI is not achieved by designing specific models to solve specific problems, but by holistic, joint models that can simultaneously solve diverse, real-world problems without too much human involvement. It is thus desirable to have new AI architectures that learn joint, fundamental representations to support a broad range of downstream AI tasks with limited additional domain knowledge, similar to what humans would do. One such proposal is XYZ-code~\cite{XYZ-code}, where monolingual text (X), audio and visual sensory signals (Y), and multilingual (Z) are organically integrated to create AI models that can speak, hear, see, and understand. Another approach is Pathways~\cite{pathways}, a single model that can generalize across millions of tasks.

\begin{figure}[t]
  \centering
   \includegraphics[width=1.0\linewidth]{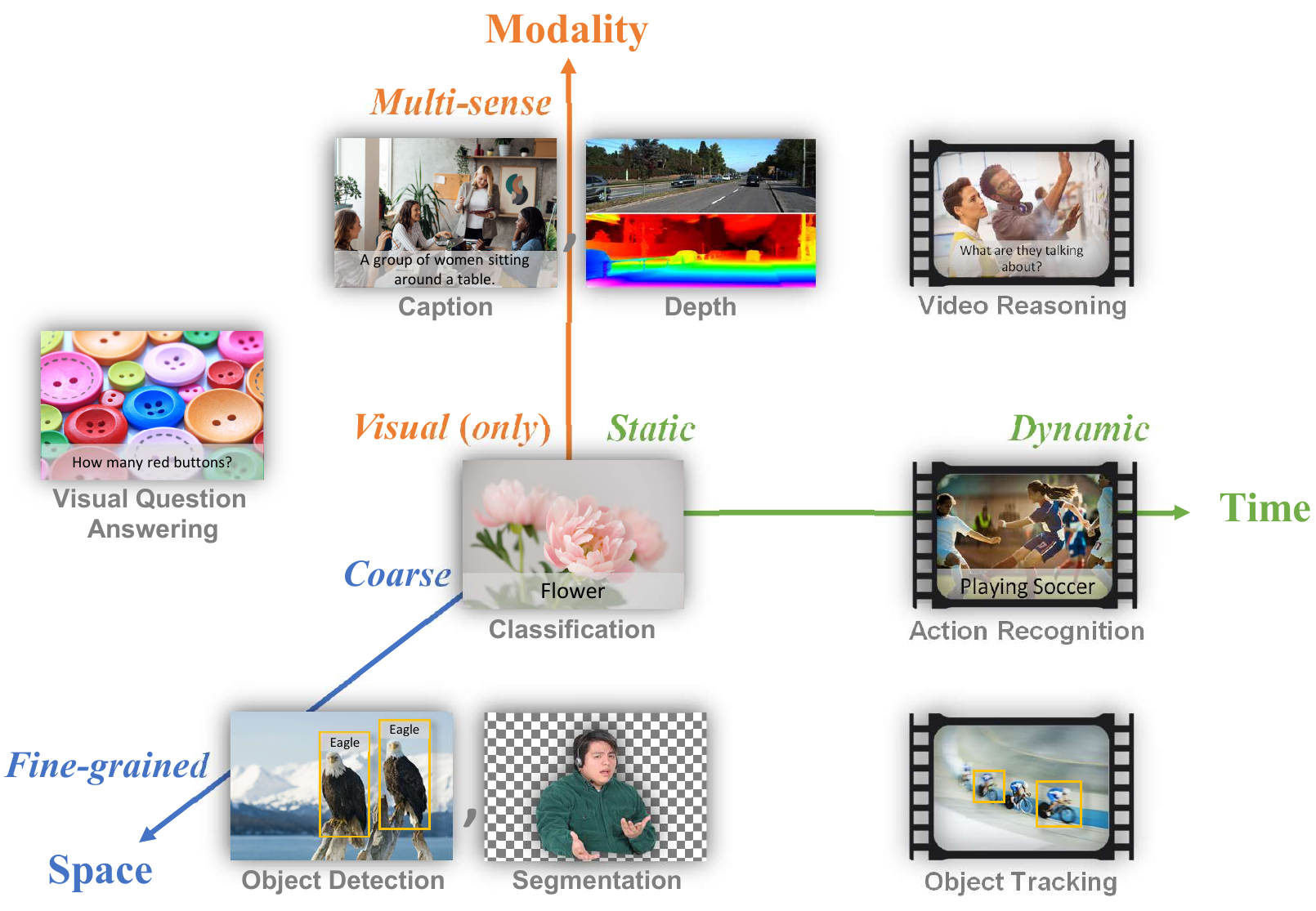}
  \caption{Common computer vision tasks are mapped to a \emph{Space-Time-Modality} space. A computer vision foundation model should serve as general purpose vision system for all of these tasks.}
  \label{fig:problem}
\end{figure}

\begin{figure*}[t]
  \centering
   \includegraphics[width=1.0\linewidth]{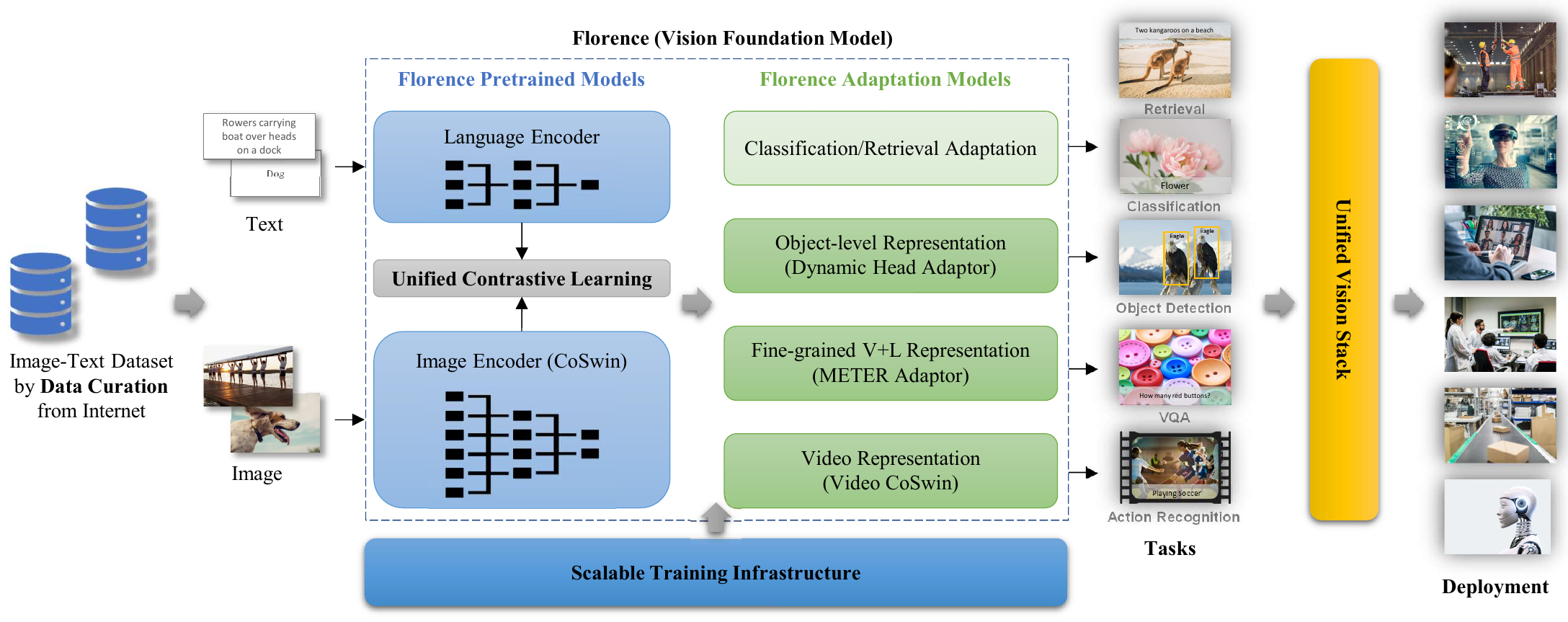}
  \caption{Overview of building Florence. Our workflow consists of data curation, unified learning,
  Transformer architectures and adaption. It shows the foundation model can be adapted to various
  downstream tasks and finally integrated into modern computer vision system to power real-world
  vision and multimedia applications. Compared with existing image-text pretraining models ~\cite{radford2021learning,jia2021scaling,Wudao2}, mainly limited on cross-modal shared representation for classification and retrieval (illustrated by \emph{light-green} adaptation module), Florence expands the representation to support object level, multiple modality, and videos respectively.}
  \label{fig:overview}
\end{figure*}

A concrete step towards this direction is the development of \emph{foundation} models. The term of \emph{foundation} model was first introduced in~\cite{bommasani2021opportunities} to refer to any model that is trained from broad data at scale that is capable of being adapted (\eg fine-tuned) to a wide range of downstream tasks. Foundation models become promising due to their impressive performance and generalization capabilities. They are quickly integrated and deployed into real-world AI systems by many researchers and developers.

Although foundation models have already demonstrated huge impact in NLP, \eg, BERT~\cite{devlin2019bert}, GPT-3~\cite{brown2020language}, in computer vision it is still standard practice to pre-train models on labeled data sets such as ImageNet~\cite{deng2009imagenet}. More recently, large-scale pre-training methods such as CLIP~\cite{radford2021learning}, ALIGN~\cite{jia2021scaling}, and Wu Dao 2.0~\cite{Wudao2}, which learn directly from Web-scale image-text pairs, show very encouraging progress for efficient transfer learning, and zero-shot capability. However, such models are restricted to image to text mapping only tasks such as classification, retrieval, and tagging.

We raise the question: ``\emph{What is the foundation model for computer vision}?". But first, in order to better define what ``foundation'' means in computer vision, we capture the spectrum of tasks in a problem space  (Figure~\ref{fig:problem}) with three orthogonal axes: {1) \emph{Space}:} from coarse (\eg ~scene-level classification) to fine-grained (\eg ~object detection), {2) \emph{Time}:} from static (\eg ~images) to dynamic (\eg ~videos), and {3) \emph{Modality}:} from RGB only to multiple senses (\eg ~captioning and depth). Due to the diversity nature of visual understanding, we redefine {\bf{\emph{foundation models for computer vision}}} to be \emph{a pre-trained model and its adapters} for solving all vision tasks in this Space-Time-Modality space, with transferability such as zero-/few-shot learning and fully fine tuning, etc. The adaptation for transferability is restricted to minimum customization for the pre-trained foundation models, such as continuing training, few epochs or few layers for fine tuning without significantly increasing or changing model parameters.

In this paper, we present an emerging paradigm for building a \emph{vision foundation model}, called \emph{Florence}. We use the name of \emph{Florence} as the origin of the trail for exploring \emph{vision foundation} models, as well as the birthplace of Renaissance. \emph{Florence} is trained on noisy Web-scale data end-to-end with a unifying objective, allowing the model to achieve best-in-class performance across a wide range of benchmarks.

The ecosystem of constructing \emph{Florence} consists of \emph{data curation}, \emph{model pretraining}, \emph{task adaptations} and \emph{training infrascturue}, as shown in Figure~\ref{fig:overview}.
\begin{itemize}
    \item \textbf{Data curation}. Diverse, large-scale data is the lifeblood of foundation models. Enabled by large amounts of publicly available images on the Internet with natural language weak supervision, we curate a new dataset of $900$ million image-text pairs for training. As Web-crawled data is usually noisy free-form texts (\eg, word, phrase or sentence), to attain more effective learning, we consider {\emph{UniCL}}, a unified image-text contrastive learning objective recently proposed in~\cite{Jianwei_UNICL2022}, which has demonstrated improvements over contrastive and supervised learning approaches.

    \item \textbf{Model pretraining} (representation learning). To learn a good representation from image-text pairs, we used a two-tower \emph{architecture} including an image encoder and a language encoder, as commonly used in CLIP~\cite{radford2021learning} and ALIGN~\cite{jia2021scaling}. For the image encoder, we chose hierarchical Vision Transformers (\eg, Swin~\cite{liu2021Swin}, CvT~\cite{Wu_2021_ICCV}, Vision Longformer~\cite{zhang2021multi}, Focal Transformer~\cite{yang2021focal}, and CSwin~\cite{dong2021cswin}). While inheriting performance benefits of the transformer self-attention operations~\cite{dosovitskiy2021image}, these hierarchical architectures model the scale invariance nature of images and have linear computational complexity with respect to image size, a property that is essential to dense prediction tasks such as object detection and segmentation.

    \item \textbf{Task adaptations}. As we have defined computer vision foundation models to adapt to various downstream tasks, it is vital for \Florence to be \emph{extensible} and \emph{transferable} for this purpose. We extended the learned feature representation along space (from scene to objects) using the dynamic head adapter ~\cite{Dai_2021_CVPR}, time (from static image to videos) via proposed video CoSwin adapter, and modality (from images to language) via METER adapter~\cite{dou2021empirical}. \Florence is designed to effectively adapted in the open world via few-shot and zero-shot transfer learning, with the ability of efficient deployment by extra training with few epochs (\eg ~in retrieval). Our model can be customized for various domains that application-developers can use.

    \item \textbf{Training infrastructure}. For both energy and cost concerns, it is critical to build foundation models with as low cost as possible. We developed scalable training infrastructure to improve training efficiency. It consists of several key techniques such as ZeRO~\cite{DBLP:journals/corr/abs-1910-02054}, activation checkpointing, mixed-precision training, gradient cache~\cite{gao2021scaling} to greatly reduce the memory consumption and thus improves the training throughput.

\end{itemize}

\Florence significantly outperforms previous large-scale pre-training methods and achieves new state-of-the-art results on a wide range of vision and vision-language benchmarks. It showed strength in zero-shot transfer in $12$ classification downstream tasks (win $9/12$, SOTA in ImageNet-1K zero-shot with top-1 accuracy of {${83.74}$} and the top-5 accuracy of {${97.18}$}), linear probe in $11$ classification downstream tasks (win $9/11$), image retrieval zero-shot ({${90.9}/ {76.7}$} R$@1$ on Flickr30K image-to-text / text-to-image, {${64.7}/ {47.2}$} R$@1$ on MSCOCO image-to-text / text-to-image) and fine-tuning ({${97.2}/ {87.9}$} R$@1$ on Flickr30K image-to-text / text-to-image, {${81.8}/ {63.2}$} R$@ 1$ on MSCOCO image-to-text/ text-to-image), object detection ({${62.4}$} mAP on COCO, {${39.3}$} mAP on Object365, {${16.2}$} AP50 on Visual Genome), VQA ({${80.36}$}), text-to-video retrieval zero-shot ({${37.6}$} R$@1$ on MSR-VTT), and video action recognition (top-1 accuracy {${86.5}/ {87.8}$} on Kinetics-400 / Kinetics-600).

\section{Approach}

\subsection{Dataset Curation}

We leverage large quantities of image-text data available publicly on the internet. Specifically,
we construct a $900$ million image-text-pair dataset, called FLD-900M (FLD stands for {\bf{FL}}orence{\bf{D}}ataset), using a programmatic data curation pipeline that processes around $3$ billion Internet images and their raw descriptions in parallel. Selection and post-filtering is employed to ensure data relevance
and quality while respecting legal and ethical constraints. To improve data quality, we
performed rigorous data filtering, similar to ALIGN~\cite{jia2021scaling}, including a simple
hash-based near-duplicate image removal, small-size image removal, image-text relevance, etc. In addition, we follow the sampling strategy introduced in~\cite{radford2021learning, ramesh2021zeroshot} with the
goal of achieving improved balance, informativeness, and learnability of the sampled dataset. The final form of the FLD-900M dataset consists of $900M$ images with $900M$ free-form texts (ranging from one word, phase to sentences), $9.7M$ unique queries, and $7.5B$ tokens in total.

\subsection{Unified Image-Text Contrastive Learning}
\label{sect:UniC}

CLIP~\cite{radford2021learning} implicitly assumes that each image-text pair has its unique caption, which allows other captions to be considered negative examples. However, in web-scale data, multiple images can be associated with identical captions. For example, in FLD-900M, there are $350M$ image-text pairs where there are more than one images corresponding to one identical text, and all images associated with the same text can be treated as positive pairs in contrastive learning.

To address this issue, we utilize a unified image-text contrastive learning
(\emph{UniCL})~\cite{Jianwei_UNICL2022}, where \emph{Florence} is pre-trained in an
image-label-description space. Given an image-text pair, we generate a triplet
$(\boldsymbol{x}, \boldsymbol{t}, \boldsymbol{y})$ via a text hash-table, where $\boldsymbol{x}$ is
the image, $\boldsymbol{t}$ is the language description (\ie, hash value), and $\boldsymbol{y}$ is
the language label (\ie, hash key) indicating the index of unique language description in the
dataset. Note that we only map identical language description to the same hash key, \ie, language
label. Thus, all image-text pairs mapped to the same label $\boldsymbol{y}$ are regarded as positive
in our universal image-text contrastive learning. Others are still regarded as negative. The unified
learning objective in the common image-label-description space unifies two popular learning
paradigms -- mapping images to the label for learning discriminative representations (\ie,
supervised learning) and assigning each description with a unique label for language-image
pre-training (\ie, contrastive learning).

Our empirical experiments indicate that long language descriptions with rich content would be more
beneficial for image-text representation learning than short descriptions (\eg, one or two words).
We have to enrich the short description by generating prompt templates such as ``A photo of the
\texttt{[WORD]}", ``A cropped photo of \texttt{[WORD]}", as data augmentation. During
training, we randomly select one template to generate $\boldsymbol{t}$ for each short language
description.

Following UniCL~\cite{Jianwei_UNICL2022}, we denote $f_{\theta}$ and $f_{\phi}$ as the image encoder and text encoder, respectively.
$\boldsymbol{u}$ and $\boldsymbol{v}$ are the normalized visual feature vector and language feature
vector, respectively, where $\boldsymbol{u} =
\frac{f_{\theta}(\boldsymbol{x})}{\|f_{\theta}(\boldsymbol{x})\|}$, and $\boldsymbol{v} =
\frac{f_{\phi}(\boldsymbol{t})}{\|f_{\phi}(\boldsymbol{t})\|}$. $\tau$ is a learnable temperature.
Given a mini-batch $\mathcal{B}$, we use a bi-directional supervised contrastive learning objective
between images and language descriptions to train the model as:
\begin{equation}\label{eq:obj_bicon}
	\mathcal{L} 	=  \mathcal{L}_{i2t} + \mathcal{L}_{t2i}.
\end{equation}
This objective contains two contrastive terms: the supervised image-to-language contrastive loss
\begin{align}\label{eq:obj_i2t_label}
\mathcal{L}_{i2t}	= & - \sum_{ i \in \mathcal{B} } \frac{1}{ |\mathcal{P}(i)|  }  \sum_{ k \in
\mathcal{P}(i) }
\log \frac{ \exp(\tau\boldsymbol{u}_{i} \boldsymbol{v}_k )  }{\sum_{ j \in \mathcal{B}}
\exp(\tau\boldsymbol{u}_{i} \boldsymbol{v}_{j} )  },
\end{align}
where $k \in \mathcal{P}(i) = \{ k | k \in \mathcal{B}, y_k = y_i\}$, and the supervised
language-to-image contrastive loss
\begin{align}\label{eq:obj_t2i_label}
\mathcal{L}_{t2i}	= & - \sum_{ j \in \mathcal{B} } \frac{1}{ |\mathcal{Q}(j)|  }  \sum_{ k \in
\mathcal{Q}(j) }
\log \frac{ \exp(\tau\boldsymbol{u}_{k} \boldsymbol{v}_j )  }{\sum_{ i \in \mathcal{B}}
\exp(\tau\boldsymbol{u}_{i} \boldsymbol{v}_{j} )  },
\end{align}
where $k \in \mathcal{Q}(j) = \{ k | k \in \mathcal{B}, y_k = y_j\} $.

The generated language prompt is not a precise description of an image, typically not as informative as the associated text descriptions from the Internet. Although including generated language prompt might not affect classification accuracy, it hurts the performance in retrieval and vision-language tasks. To mitigate the negative effect from augmented prompts, our training is separated into two stages. In the first stage, we use
all data including augmented texts for training; while in the second stage, we exclude all augmented
data for continuing training. We trained $1M$ iterations in the first stage, and continuously
trained $180K$ iterations in the second stage. The Adam optimizer with decoupled weight decay
regularization is utilized for model training. The image size is $224\times224$ and the maximum
language description length is truncated at $76$. The batch size is $24,576$. We further trained
$80K$ iterations at a higher resolution of $384\times384$ to boost the performance, which follows
existing pre-training approaches.

\subsection{Transformer-based Florence Pretrained Models}
\label{sect:architecture}

Our \emph{Florence} pretrained model uses a two-tower architecture: a 12-layer transformer~\cite{NIPS2017_3f5ee243} 
as language encoder, similar to CLIP~\cite{radford2021learning}, and a
hierarchical Vision Transformer as the image encoder. The hierarchical Vision Transformer is
a modified Swin Transformer~\cite{liu2021Swin} with convolutional embedding, called \emph{CoSwin} Transformer. Specifically, we replace the patch embedding and patch merging
modules in the Swin Transformer~\cite{liu2021Swin} with the convolutional embedding layers as
described in CvT~\cite{Wu_2021_ICCV}. We use the \emph{CoSwin} Transformer with global average
pooling to extract image features. Two linear projection layers are added on top of the image
encoder and language encoder to match the dimensions of image and language features. Our \emph{Florence} pretrained model has
in total $893M$ parameters, including the language transformer with $256M$ parameters and the
\emph{CoSwin}-H transformer with $637M$ parameters. The model takes $10$ days to train on $512$
NVIDIA-A100 GPUs with 40GB memory per GPU.

\subsection{Object-level Visual Representation Learning}

We extend the \emph{Florence} pretrained model to learn fine-grained (\ie, object-level) representation, which is fundamental to
dense prediction tasks such as object detection. For this goal, we add an adaptor \emph{Dynamic Head}~\cite{Dai_2021_CVPR} (or Dynamic DETR~\cite{Dai_2021_ICCV}), a unified attention
mechanism for the detection head, to the pretrained image encoder (\ie, \emph{CoSwin}). We can continue visual representation learning from coarse (scene) to fine (object).

\begin{figure}[t]
  \centering
   \includegraphics[width=1.0\linewidth]{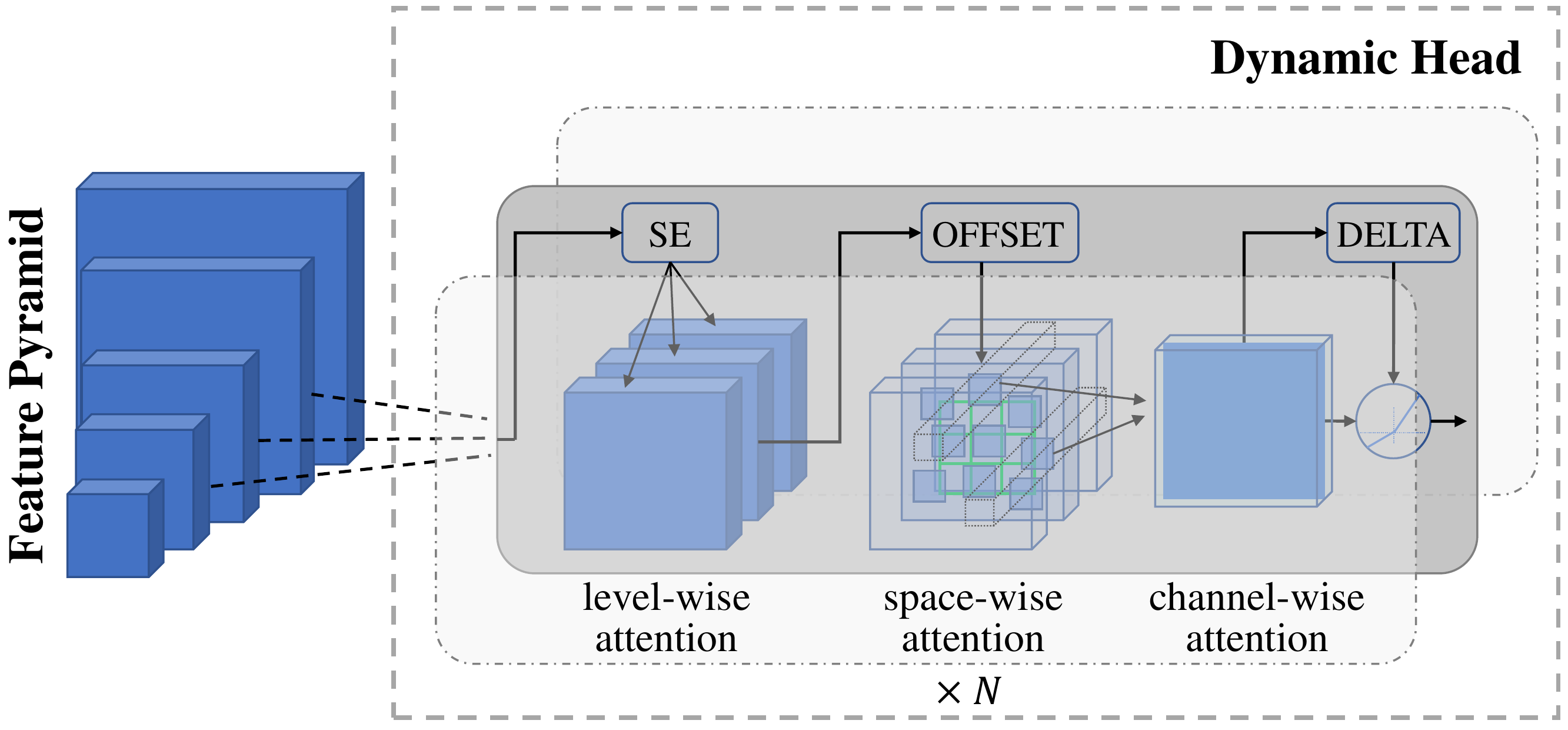}
  \caption{Dynamic Head~\cite{Dai_2021_CVPR} adapter is used for object-level visual representation learning.}
  \label{fig:dyhead}
\end{figure}

Based on the hierarchical structure of the image encoder \emph{CoSwin}-H, we can get the output
feature pyramids from the different scale levels. The feature pyramid scale levels can be concatenated and
scaled-down or scaled-up into a 3-dimensional tensor with dimensions $level \times space \times
channel$. The key idea of \emph{Dynamic Head}~\cite{Dai_2021_CVPR} is to deploy three attention
mechanisms, each on one of the orthogonal dimensions of the tensor, \ie, level-wise, spatial-wise, and
channel-wise. Compared with building a single self-attention mechanism over this tensor, \emph{Dynamic
Head} makes the computation more affordable and enables more efficient learning. The above three
attention mechanisms are applied sequentially, and we can effectively stack multiple blocks
consisting of such three attention layers together. Figure~\ref{fig:dyhead} shows the \emph{Dynamic
Head} building blocks. In this work, \emph{Dynamic Head} is trained with the one-stage ATSS
framework and losses.

We have constructed a large-scale object detection dataset, called FLOD-9M (for {\bf{FL}}orence {\bf{O}}bject detection {\bf{D}}ataset), for object detection
pre-training. We merge several well-known object detection datasets, including
COCO~\cite{lin2015microsoft}, LVIS~\cite{Gupta_2019_CVPR}, OpenImages~\cite{openimages},
Object365~\cite{Shao_2019_ICCV}. In addition, we generate pseudo bounding boxes on ImageNet-22K
dataset~\cite{deng2009imagenet} by following~\cite{ZophGLCLC020_nips}, which further enlarges our
data. In the end, FLOD-9M consists of $8,967,286$ images, $25,190$ object categories, and
$33,408,237$ bounding boxes including annotations and pseudo labels. We then pre-train our
\emph{Dynamic Head} model for $12$ epochs with batch size $128$, which takes $7$ days on $128$
NVIDIA-A100 GPUs.

\subsection{Fine-Grained V+L Representation Learning}

We use METER~\cite{dou2021empirical} adapter to expand to fine-grained vision-language representation. In the vision-language area, \eg ~visual question answering (VQA) and image captioning, fine-grained representation (\ie, object-level) is indispensable. Thus, the object detector has been a de facto tool for image feature extraction,
followed by a fusion network for prediction in many works ~\cite{00010BT0GZ18,li:oscar,Zhang_2021_CVPR,abs-2012-06946,abs-2104-02096,chen:uniter}.
Recently, there is an increasing trend~\cite{abs-2104-03135,xue2021probing,wang2021simvlm,KimSK21,dou2021empirical} of end-to-end approaches to reduce dependency on the object bounding box, which instead consider grid-based feature representations as the fine-grained features for V+L tasks.

\begin{figure}[t!]
    \centering
    \includegraphics[width=1.0\linewidth]{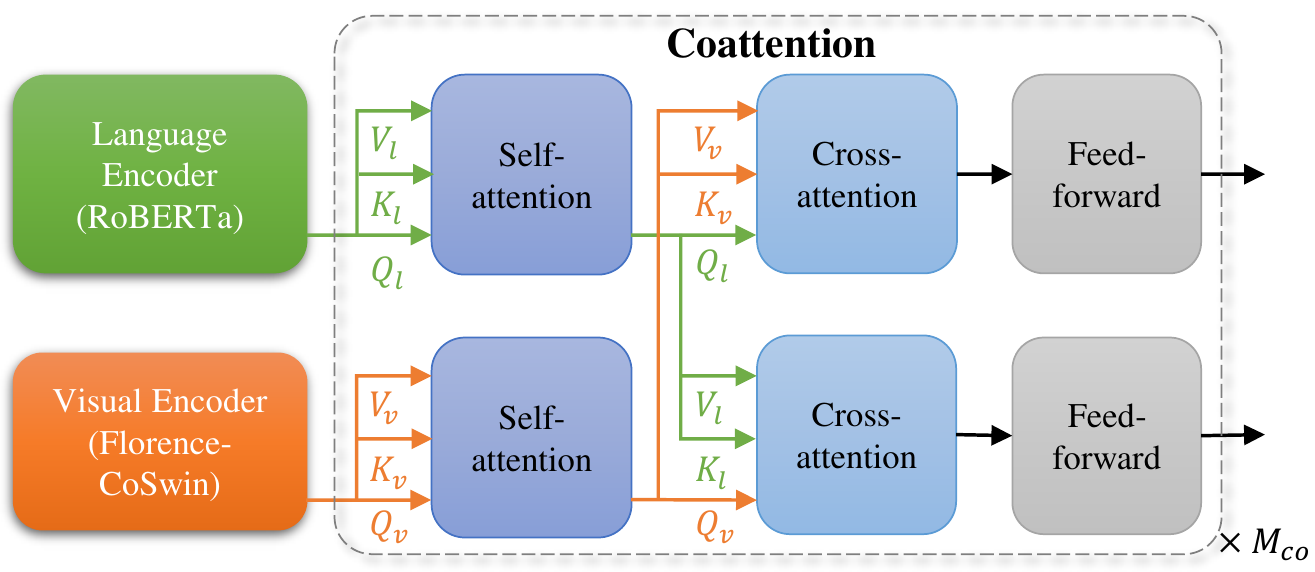}
    \caption{METER~\cite{dou2021empirical} is used as Florence V+L adaptation model, trained with
    the image-text matching (ITM) loss
    and the masked language modeling (MLM) loss.}
    \label{fig:florence_meter}
\end{figure}

In the \emph{Florence} V+L adaptation model, we replace the image encoder of METER~\cite{dou2021empirical} with \emph{Florence} pretrained model \emph{CoSwin}, and use a pretrained  Roberta~\cite{liu2019roberta} as the language encoder, shown in Figure~\ref{fig:florence_meter}. The \emph{Florence} pretrained language encoder can be used for this adapter as it utilizes BERT-based architecture. Then, the two modalities are fused together to learn the contextual representation with a transformer network based on co-attention. The co-attention model
(Figure~\ref{fig:florence_meter}) allows feeding the text and visual features to two $M_{co}$-layer transformers separately, and each top transformer encoding layer consists of one self-attention block, one cross-attention block, and one feed-forward network block. We first train the model with the image-text matching loss and the masked-language modeling loss. Then, we fine-tune the model on the downstream task, such as VQA~\cite{GoyalKSBP16} task.

\subsection{Adaption to Video Recognition}
\label{sect:adapt_video}

The self-attention based design in Transformer makes it possible to unify the systems of image and video recognition. Our Video \emph{CoSwin} adapter can borrow the image encoder from \emph{CoSwin} for the video domain with minimum changes, similar to prior work~\cite{liu2021video}. First, the image tokenization layer is replaced with a video tokenization layer. Accordingly, video \emph{CoSwin} replaces the tokenization layer of \emph{CoSwin} (in Section~\ref{sect:architecture}) from 2D convolutional layers to 3D convolutional layers, which converts each 3D tube into one token. As the initialization to 3D convolutional weights, the pre-trained 2D convolutional weights of \emph{CoSwin} are duplicated along the temporal dimension and divided by the temporal kernel size to keep the mean and
variance of the output unchanged. Second,  video \emph{CoSwin} uses the 3D convolution-based patch merging operator instead of the 2D patch merging operator used in~\cite{liu2021video}. Such overlapped token merging
can enhance spatial and temporal interactions among tokens. Third, we follow prior work~\cite{liu2021video} to replace the 2D shifted window design with 3D shifted local windows in self-attention layers. We duplicate the 2D relative positional embedding matrix from the pre-trained \emph{CoSwin} along the temporal dimension to initialize the 3D positional embedding matrix. In this way, the 2D relative positional embedding is the same for each temporal shift. In addition, all other layers and weights
(including self-attention, FFN) can be inherited directly from the pre-trained \emph{CoSwin}. To mitigate memory issues in the video training, we adopt the dynamic window size strategy, i.e., a relatively small window size in early stages of \emph{CoSwin}, and large window sizes in its later stages.

\begin{table*}[ht]
\centering
\setlength{\tabcolsep}{7.1pt}
\small
\renewcommand{\arraystretch}{1.35}
\begin{tabular}{l|cccccccccccc}
\toprule
 & \rotatebox{90}{Food101} & \rotatebox{90}{CIFAR10} & \rotatebox{90}{CIFAR100} &
 \rotatebox{90}{SUN397} & \rotatebox{90}{Stanford Cars} & \rotatebox{90}{FGVC Aircraft} &
 \rotatebox{90}{VOC2007} & \rotatebox{90}{DTD} & \rotatebox{90}{Oxford Pets} &
 \rotatebox{90}{Caltech101} & \rotatebox{90}{Flowers102} & \rotatebox{90}{ImageNet}\\ \midrule
CLIP-ResNet-50x64 & 91.8 & 86.8 & 61.3 & 48.9 & 76.0 & 35.6 & 83.8 & 53.4
& 93.4 & 90.6 & 77.3 & 73.6 \\
CLIP-ViT-L/14 ($@ 336$pix) & 93.8 & \underline{\bf{95.7}} & 77.5 & 68.4 &
78.8 & 37.2 & 84.3 & 55.7 & 93.5 & 92.8 & 78.3 & 76.2 \\
FLIP-ViT-L/14 & 92.2 & \underline{\bf{95.7}} & 75.3 & 73.1 & 70.8 &
\underline{\bf{60.2}} & - & 60.7 & 92.0 & 93.0 & \underline{\bf{90.1}} & 78.3 \\
Florence-CoSwin-H ($@ 384$pix) & \underline{\bf{95.1}} & 94.6 & \underline{\bf{77.6}} &
\underline{\bf{77.0}} & \underline{\bf{93.2}} & 55.5 & \underline{\bf{85.5}} & \underline{\bf{66.4}}
& \underline{\bf{95.9}} & \underline{\bf{94.7}} & 86.2 & \underline{\bf{83.7}} \\ \bottomrule
\end{tabular}
\caption{Zero-shot transfer of image classification comparisons on 12 datasets: CLIP-ResNet-50x64~\cite{radford2021learning}, FLIP-ViT-L/14~\cite{yao2021filip}.}
\label{tab:zero-shot}
\end{table*}

\subsection{Scalable Training Infrastructure}

\begin{figure}[t!]
    \centering
    \includegraphics[width=1.0\linewidth]{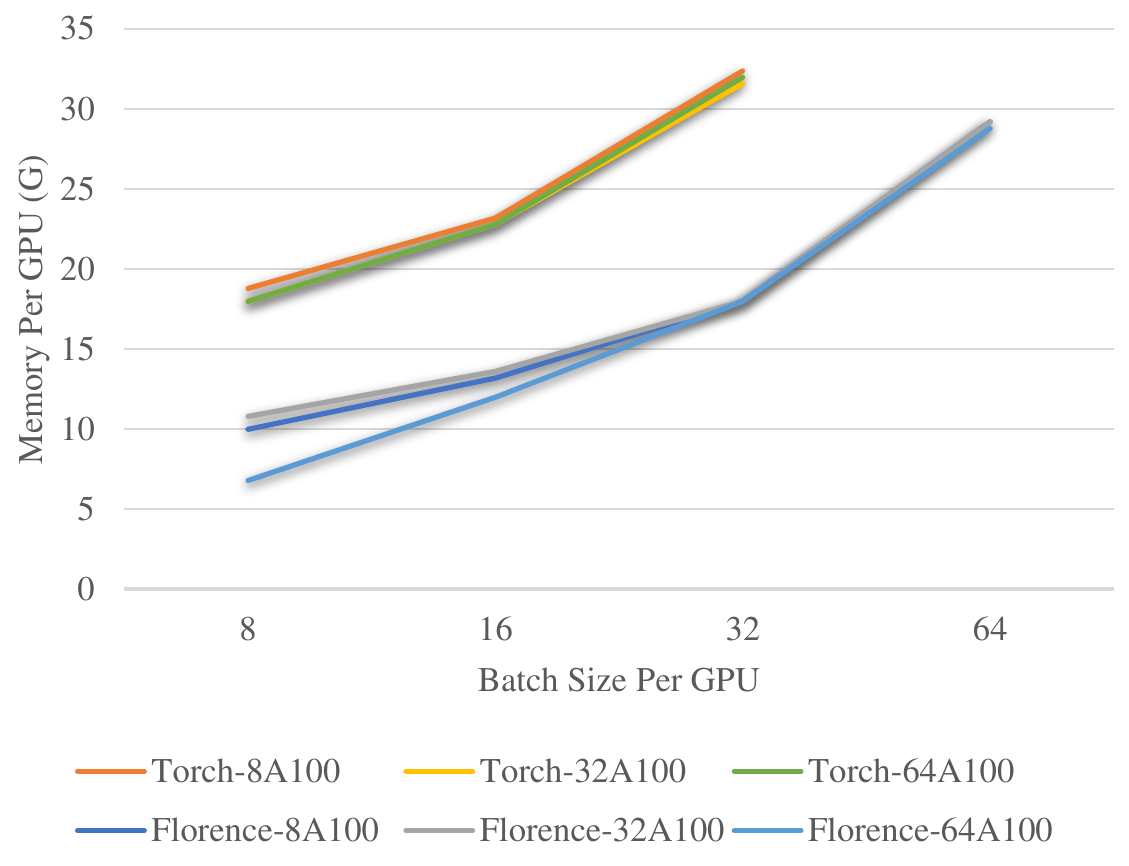} \vspace{-0.7em}
    \caption{GPU memory reduction for various batch sizes. We compared the profiling between Torch (w/o optimization) and Florence (w/ optimization) on various number of GPUs.}
    \label{fig:florence_profiling}
\end{figure}

To train the \emph{Florence} model on our large-scale dataset, our scalable training infrastructure
faces two main challenges: reducing memory cost on each GPU and increasing the throughput. Reducing
the memory cost allows us to feed more data into each GPU and use a larger batch size, which has
been proved to be effective for contrastive learning.
Increasing the throughput can significantly speed up the whole training process and thus reduce
carbon emissions. We have developed several techniques that can be combined to achieve the two
goals:

\begin{description}
  \item[Zero Redundancy Optimizer (ZeRO)] The ZeRO
      technique~\cite{DBLP:journals/corr/abs-1910-02054} partitions the optimizer states,
      gradients and parameters across the GPUs and each partition is only updated locally. Thus,
      the memory consumption is largely reduced.
  \item[Activation Checkpointing] For a checkpointed model component, \eg, multi-head attention,
      it reruns a forward pass during backward pass. In this way, the internal gradients in the
      component do not need to be stored in the forward pass and then reduce the memory cost in
      the training.
  \item[Mixed-precision Training] In mixed-precision training, various operations are trained with
      different numerical precision (\ie, float-32 or float-16). Float-32 is used for numerically
      less stable operations, such as layer normalization; while float-16 is used for the other
      operations. Such a combination improves the training throughput and maintains the model
      performance.
  \item[Gradient Cache] The gradient cache technique~\cite{gao2021scaling} is able to increase the
      total batch size in a training step. A large batch size is shown to be beneficial to learn better
      representations in previous works. However, it is bounded by available GPU memory. To
      resolve this problem, we factor the contrastive loss by breaking the large batch gradient
      update into several sub-updates that can fit into GPU memory. It enables us to train big
      models with a large batch size.
\end{description}

Thanks to these above optimizations, we can achieve consistent improvement in reducing GPU memory for variable batch sizes on various numbers of NVIDIA-A100s, shown in Figure~\ref{fig:florence_profiling}.

\section{Experiments}

\begin{table*}[ht]
\centering
\setlength{\tabcolsep}{8.3pt}
\small
\renewcommand{\arraystretch}{1.35}
\begin{tabular}{l|ccccccccccc}
\toprule
 & \rotatebox{90}{Food101} & \rotatebox{90}{CIFAR10} & \rotatebox{90}{CIFAR100} &
 \rotatebox{90}{SUN397} & \rotatebox{90}{Stanford Cars} & \rotatebox{90}{FGVC Aircraft} &
 \rotatebox{90}{VOC2007} & \rotatebox{90}{DTD} & \rotatebox{90}{Oxford Pets} &
 \rotatebox{90}{Caltech101} & \rotatebox{90}{Flowers102} \\ \midrule
SimCLRv2-ResNet-152x3 & 83.6 & 96.8 & 84.5 & 69.1 & 68.5 & 63.1 & 86.7 & 80.5 &
92.6 & 94.9 & 96.3  \\
ViT-L/16 ($@ 384$pix) & 87.4 & 97.9 & 89.0 & 74.9 & 62.5 & 52.2 & 86.1 &
75.0 & 92.9 & 94.7 & 99.3  \\
EfficientNet-L2 ($@ 800$pix) & 92.0 & \underline{\bf{98.7}} &
\underline{\bf{89.0}} & 75.7 & 75.5 & 68.4 & 89.4 & 82.5 & 95.6 & 94.7 & 97.9  \\
CLIP-ResNet-50x64 & 94.8 & 94.1 & 78.6 & 81.1 & 90.5 & 67.7 & 88.9 & 82.0
& 94.5 & 95.4 & 98.9 \\
CLIP-ViT-L/14 ($@ 336$pix) & 95.9 & 97.9 & 87.4 & 82.2 & 91.5 & 71.6 &
89.9 & 83.0 & 95.1 & 96.0 & 99.2 \\
Florence-CoSwin-H ($@ 384$pix) & \underline{\bf{96.2}} & 97.6 & 87.1 & \underline{\bf{84.2}} &
\underline{\bf{95.7}} & \underline{\bf{83.9}} & \underline{\bf{90.5}} & \underline{\bf{86.0}} &
\underline{\bf{96.4}} & \underline{\bf{96.6}} & \underline{\bf{99.7}} \\  \bottomrule
\end{tabular}
\caption{Comparisons of image classification linear probing on 11 datasets with existing state-of-the-art models, including SimCLRv2~\cite{chen2020big}, ViT~\cite{dosovitskiy2020vit}, EfficientNet~\cite{Xie_2020_CVPR}, and CLIP~\cite{radford2021learning}.}
\label{tab:linear}
\end{table*}

\subsection{Zero-shot Transfer in Classification}
\label{sec:zero-shot}

In computer vision, zero-shot learning usually refers to the study of predicting classes that are
defined via descriptive text. As a vision foundation model, \emph{Florence} can be directly used to
predict if an image and a text snippet are semantically matched together in the task dataset. We
follow the same method of CLIP~\cite{radford2021learning} to perform zero-shot classification. For each
dataset, we use the names of all the classes in the dataset as the set of potential text pairings
and predict the most probable (image, text) pair according to \emph{Florence}. We compute the
feature embedding of the image for \emph{CoSwin} and the feature embedding of the set of possible
texts by the language encoder. The cosine similarities among these embeddings are then calculated,
and then we rank the similarity scores over all the classes to select the Top-1 or Top-5 classes as
the predicted classes. Here, we do not need to compute the normalized cosine similarity as done
in~\cite{radford2021learning}, since it won't affect the ranking order of final results.

We evaluate our \emph{Florence} model on the ImageNet-1K dataset and 11 downstream datasets from the
well-studied evaluation suit introduced by~\cite{kornblith2019better}. Note that our benchmarks
exclude the Birdsnap~\cite{6909656} dataset from 12 original classification datasets introduced
in~\cite{kornblith2019better}, because $20\%$ of the image URLs provided by the authors
are invalid. We follow the same prompt templates and engineering, and ensembling
as previously proposed in~\cite{radford2021learning} for evaluating zero-shot performance. For all zero-shot tasks in this
paper, we follow the setup in CLIP~\cite{radford2021learning} and ALIGN~\cite{jia2021scaling} to
remove near-duplicate test images from our training data. Table~\ref{tab:zero-shot} shows the
results over these 12 datasets, in comparison with the best performance achieved by both CLIP ResNet
and Vision Transformer models, and the concurrent work FILIP~\cite{yao2021filip}. \emph{Florence} outperforms on $9/12$ tasks compared with state-of-the-art methods. We achieved a remarkable improvement in the zero-shot transfer on ImageNet-1K -- the top-1 accuracy of $83.74\%$ ($+5.6\%$ over SOTA result),  and the top-5 accuracy of
$97.18\%$.

\subsection{Linear Probe in Classification}

Linear probe as another main metric for evaluating representation quality has been used in most
recent studies, including self-supervised learning~\cite{pmlr-v119-chen20j, chen2020big},
self-training with noisy student~\cite{Xie_2020_CVPR} and contrastive
learning~\cite{radford2021learning}. We follow the same setting and implementation of
CLIP~\cite{radford2021learning} for linear evaluation, where the image encoder (or vision backbone)
is frozen, and only the appended linear layers can be fine-tuned on the downstream datasets. We use public available models (shown in Table 10~\cite{radford2021learning}) to verify the correctness of our own implementation. The variance between our reproduced results and their reported results is $\pm 0.1$ for each task. Our linear evaluation considers 11 classification
benchmarks which are also used for our zero-shot transfer of classification. We compared our results with state-of-the-art methods with their best performance models, including SimCLRv2~\cite{chen2020big},
ViT~\cite{dosovitskiy2020vit}, Noisy Student~\cite{Xie_2020_CVPR} and
CLIP~\cite{radford2021learning} on Table~\ref{tab:linear}. Our results are consistently  better than
existing state-of-the-art results, expect for two datasets: CIFAR10, CIFAR100. On the two datasets, the input image
resolution is quite low (\ie, $32 \times 32$). Training with higher resolution definitely boosts the performance,such as  Efficient-L2~\cite{Xie_2020_CVPR} which achieves the best accuracy compared with all other approaches trained on lower-resolution images.

\subsection{ImageNet-1K Fine-tune Evaluation}
\label{sect:imagenet_finetune}

\emph{Florence} can be easily adapted to support continual fine-tuning on target classification
tasks. We do not change or add anything into our architecture, but continue the training on
task-specific data using the same pre-training loss (shown in Equation~\ref{eq:obj_bicon}). We feed the class name to the text
encoder of \emph{Florence} to get the text feature embedding. We use the same prompt templates as in
~\cite{radford2021learning, jia2021scaling} to expand the descriptions of
ImageNet~\cite{deng2009imagenet} class names.

We evaluate the performance of continual fine-tuning on ImageNet ILSVRC-2012
benchmark~\cite{deng2009imagenet}. Our image encoder \emph{CoSwin}-H is
fine-tuned at the resolution of $512\times512$ with a batch size of $8,192$ for $10$ epochs. We use a cosine learning rate decay scheduler with $500$ warmup steps and a peak learning rate of
$0.00002$. The comparisons with state-of-the-art results are shown in
Table~\ref{tab:imagenet_result}. Our model outperforms BiT~\cite{kolesnikov2020big} with larger model size and ALIGN~\cite{jia2021scaling} trained from more data in terms of Top-1 and Top-5 accuracy. Our result is slightly worse than SOTA~\cite{dai2021coatnet}, but their model and data scale are both $3\times$ larger.

\begin{table}[t]
\centering
\setlength{\tabcolsep}{6.3pt}
\small
\renewcommand{\arraystretch}{1.35}
\begin{tabular}{l|rrcc}
\toprule
\multirow{2}{*}{Model} & \multirow{2}{*}{Params} & \multirow{2}{*}{Data\;} &
\multicolumn{2}{c}{Accuracy} \\
& & & Top-1 & Top-5\\ \midrule
BiT-L-ResNet152x4 & 928M & 300M & 87.54 & 98.46  \\
ALIGN-Efficient-L2 & 480M & 1800M & 88.64 & 98.67  \\
ViT-G/14 & 1843M & 3000M & 90.45 & -  \\
CoAtNet-7 & 2440M & 3000M & \underline{\bf{90.88}} & - \\
Florence-CoSwin-H & 637M & 900M & 90.05 & \underline{\bf{99.02}}  \\ \bottomrule
\end{tabular}
\caption{Classification fine tuning on ImageNet-1K. Florence is compared with: BiT-L-ResNet152x4~\cite{kolesnikov2020big}, ALIGN-Efficient-L2~\cite{jia2021scaling}, ViT-G/14~\cite{zhai2021scaling}, CoAtNet-7~\cite{dai2021coatnet} in terms of model scale, data scale and Top-1/Top-5 accuracy.}
\label{tab:imagenet_result}
\end{table}

\subsection{Few-shot Cross-domain Classification}

The Cross-Domain Few-Shot learning benchmark \cite{cdfsl} is used to measure an algorithm's capability to adapt to downstream few-shot target tasks, containing domains with varying levels of dissimilarity to typical consumer photographs. The datasets in the benchmark include: CropDisease~\cite{plantdisease} (plant leaf images, 38 disease states over 14 plant species), EuroSAT~\cite{eurosat} (RGB satellite images, 10 categories), ISIC 2018~\cite{isic2018,ham10000} (dermoscopic images of skin lesions, 7 disease states), and ChestX~\cite{chestx} (Chest X-rays, 16 conditions). Exemplar image for each dataset is shown on the top of Table~\ref{tab:cdfsl}. The evaluation protocol involves 5-way classification across 5-shot, 20-shot, and 50-shot. The classes and shots are randomly sampled for each episode, for 600 episodes per way and shot. Average accuracy over all episodes is reported.

To predict the class, we append a single linear layer as an adapter head to our image encoder \emph{CoSwin}. Training occurs over 100 epochs per episode. We use SGD with momentum, with learning rate and momentum values of $0.9/ 0.0002$, respectively, for \emph{CoSwin}, and $0.99/0.01$, respectively, for the adapter head. Horizontal data flip augmentation is used for training and test, and dropout of $0.5$ is used between the image encoder and the classifier head.

Table \ref{tab:cdfsl} shows the results of adapting our model to the CD-FSL benchmark, in comparison to the winner of the challenge benchmark~\cite{cdfsltop}, which employs ensembes and transductive learning. By comparison, we employ a single model and no transduction on the test data is performed, yet we achieve higher results without any ``bells and whistles''.

\begin{table}[t]
\centering
\begin{minipage}[t]{1.0\linewidth}
  \centering
\includegraphics[width=1.0\linewidth]{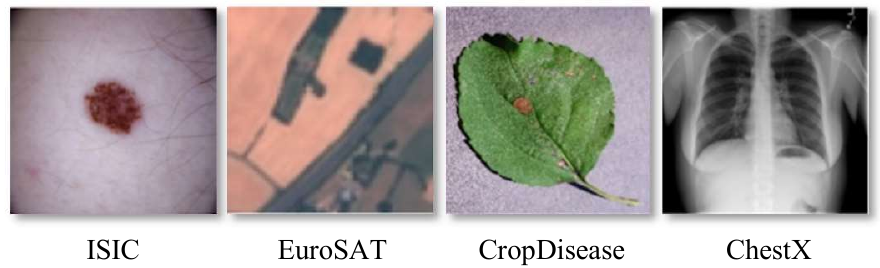}
\end{minipage}\vspace{0.3em}
\setlength{\tabcolsep}{3.8pt}
\small
\renewcommand{\arraystretch}{1.35}
\begin{tabular}{r|l|cccc|c}
\toprule
 & Model & ISIC & EuroSAT & CropD & ChestX & mean \\ \midrule
\multirow{2}{*}{5-shot} & CW & 57.4 & 88.1 & 96.6 & 29.7 & 68.0 \\
 & Florence & 57.1 & 90.0 & 97.7 & 29.3 & \underline{\bf{68.5}} \\ \midrule
\multirow{2}{*}{20-shot} & CW & 68.1 & 94.7 & 99.2 & 38.3 & 75.1 \\
 & Florence & 72.9 & 95.8 & 99.3 & 37.5 & \underline{\bf{76.4}} \\ \midrule
\multirow{2}{*}{50-shot} & CW & 74.1 & 96.9 & 99.7 & 44.4 & 78.8 \\
 & Florence & 78.3 & 97.1 & 99.6 & 42.8 & \underline{\bf{79.5}} \\ \bottomrule
\end{tabular}
\caption{Comparison with CW~\cite{cdfsltop} (CD-FSL Challenge 2020 Winner) on CD-FSL benchmark. The average result comparison is $74.8$ (Florence) vs. $73.9$ (CW).}
\label{tab:cdfsl}
\end{table}

\subsection{Image-Text Retrieval}

\begin{table*}[t]
\centering
\setlength{\tabcolsep}{9.5pt}
\small
\renewcommand{\arraystretch}{1.35}
\begin{tabular}{cl|cccc|cccc}
\toprule
& & \multicolumn{4}{c|}{Flickr30K ($1K$ test set)} & \multicolumn{4}{c}{MSCOCO ($5K$ test set)} \\
& Method & \multicolumn{2}{c}{Image $\to$ Text} & \multicolumn{2}{c|}{Text $\to$ Image} &
\multicolumn{2}{c}{Image $\to$ Text} & \multicolumn{2}{c}{Text $\to$ Image} \\
& & R@1 & R@5  & R@1 & R@5  & R@1 & R@5  & R@1 & R@5 \\
\midrule
\multirow{5}{*}{\it{Zero-shot}} & ImageBERT~\cite{qi:imagebert} & 70.7 & 90.2 & 54.3 & 79.6 & 44.0 &
71.2 & 32.3 & 59.0  \\
& UNITER~\cite{chen:uniter} & 83.6 & 95.7 & 68.7 & 89.2 & - & - & - & - \\
& CLIP~\cite{radford2021learning} & 88.0 & 98.7 & 68.7 & 90.6 & 58.4 & 81.5 & 37.8 & 62.4 \\
& ALIGN~\cite{jia2021scaling} & 88.6 & 98.7  & 75.7 & \underline{\bf{93.8}}  & 58.6 & 83.0  & 45.6 &
69.8  \\
& FLIP~\cite{yao2021filip} & 89.8 & \underline{\bf{99.2}}  & 75.0 & 93.4  & 61.3 & 84.3  & 45.9 &
70.6  \\
& Florence & \underline{\bf{90.9}} & \underline{\bf{99.1}} &  \underline{\bf{76.7}} &
\underline{\bf{93.6}} & \underline{\bf{64.7}} & \underline{\bf{85.9}} & \underline{\bf{47.2}} &
\underline{\bf{71.4}} \\
\hline
\multirow{7}{*}{\it{Fine-tuned}} & GPO~\cite{chen:vsepooling} & 88.7 & 98.9  & 76.1 & 94.5& 68.1 &
90.2 & 52.7 & 80.2  \\
& UNITER~\cite{chen:uniter} & 87.3 & 98.0 & 75.6 & 94.1 & 65.7 & 88.6 &  52.9 & 79.9  \\
& ERNIE-ViL~\cite{yu:ernie-vil} & 88.1 & 98.0 & 76.7 & 93.6   & - & - & - & - \\
& VILLA~\cite{gan:villa} & 87.9 & 97.5 & 76.3 & 94.2 & - & - & - & - \\
& Oscar~\cite{li:oscar} & - & - & - & - & 73.5 & 92.2 & 57.5 & 82.8 \\
& ALIGN~\cite{jia2021scaling} & {95.3} & {99.8} & {84.9} & {97.4} & {77.0} & {93.5} & {59.9} &
{83.3} \\
& FLIP~\cite{yao2021filip} & 96.6 & \underline{\bf{100.0}}  & 87.1 & 97.7  & 78.9 & 94.4  & 61.2 &
84.3  \\
& Florence & \underline{\bf{97.2}} & \underline{\bf{99.9}} & \underline{\bf{87.9}} &
\underline{\bf{98.1}} & \underline{\bf{81.8}} & \underline{\bf{95.2}}& \underline{\bf{63.2}} &
\underline{\bf{85.7}} \\
\bottomrule
\end{tabular}
\caption{Image-text retrieval comparisons on Flickr30K and MSCOCO datasets (zero-shot and
fine-tuned).}
\label{tab:flickr30k_mscoco_result}
\end{table*}

Table~\ref{tab:flickr30k_mscoco_result} presents the zero-shot transfer and fine-tuning performance
of \emph{Florence} for both text and image retrieval on the Flickr30k~\cite{plummer2016flickr30k}
and MSCOCO~\cite{lin2015microsoft} datasets.

For zero-shot retrieval, we feed the input text (or image) to the language (or image) encoder of
\emph{Florence} to get the feature embeddings, and also compute the feature embeddings of the set of
possible images (or texts) by the image (or language) encoder. Then we compute cosine similarity of
these embeddings and rank the similarity scores over the testing set to select the Top-1 or Top-5
results. Zero-shot \emph{Florence} matches or outperforms all prior zero-shot results on these two
datasets.

For fine-tuning retrieval, we continuously train our language and text encoders on the target
image-text pair data, as well as classification fine-tuning (shown in
Section~\ref{sect:imagenet_finetune}). We fine-tune our model with a batch size of $3,072$ for $12$
epochs. We use the cosine learning rate decay scheduler with $200$ warmup steps and a peak learning
rate of $0.00002$. Our results are superior to all previous fine-tuning results on the two
datasets. Moreover, our fine tuning on retrieval is more efficient, with only roughly $6\%$ and $8\%$ fine-tuning epochs of ALIGN~\cite{jia2021scaling} on Flickr30k and MSCOCO respectively.

\subsection{Object Detection and Zero-shot Transfer}

Object detection is one of the most prominent applications in computer vision. Compared with
existing large-scale pre-trained models (\eg, CLIP~\cite{radford2021learning},
ALIGN~\cite{jia2021scaling}, Wu Dao 2.0~\cite{Wudao2}), \emph{Florence} is more desirable for object
detection since its adaptation helps learn visual representation at the object level.  We evaluate
its performance of object-level visual representations via fine-tuned object
detection and zero-shot transfer tasks.

\paragraph{Fine-tuning} We evaluate fine-tuning on three popular object detection datasets:
COCO~\cite{lin2015microsoft}, Object365~\cite{Shao_2019_ICCV}, and Visual
Genome~\cite{krishnavisualgenome}. For COCO, we increase the maximum image side to $2,500$ and
fine-tune with multi-scale training for $12$ epochs. We follow the same multi-scale testing strategy
widely used in existing state-of-the-art approaches. For Object365, we use the same input resolution
of images (\ie, the maximum image side $1,333$) as the Multi-dataset Detection\footnote{This work was ranked 1-st
in the object detection track of ECCV 2020 Robust Vision Challenge.}~\cite{zhou2021simple} for
fine-tuning. For Visual Genome, we increase the maximum side of input resolution to $3,000$ and
fine-tune with multi-scale training for $24$ epochs. To leverage attributes annotations in Visual
Genome, we insert an $1\times1$ ROI pool on the final stage of \emph{CoSwin} backbone to extract
features for attribute learning, which allows the object detection adapter being optimized for multi-task
learning.

\begin{table}[t]
\centering
\setlength{\tabcolsep}{12.5pt}
\small
\renewcommand{\arraystretch}{1.35}
\begin{tabular}{c|l|c}
\toprule
Benchmark & Model & AP  \\ \midrule
\multirow{3}{*}{\it{COCO miniVal}} & DyHead & 60.3 \\
& Soft Teacher & 60.7 \\
& Florence & \underline{\bf{62.0}} \\ \midrule
\multirow{3}{*}{\it{COCO test-Dev}} & DyHead & 60.6 \\
& Soft Teacher & 61.3 \\
& Florence & \underline{\bf{62.4}} \\ \midrule
\multirow{2}{*}{\it{Object365}} & Multi-dataset Detection & 33.7 \\
& Florence & \underline{\bf{39.3}} \\ \midrule
\multirow{2}{*}{\it{Visual Genome}} & VinVL & 13.8 \\
& Florence & \underline{\bf{16.2}} \\
\bottomrule
\end{tabular}
\caption{Object detection fine tuning comparisons with state-of-the-art methods, including DyHead~\cite{Dai_2021_CVPR}, Soft Teacher~\cite{Xu_2021_ICCV}, Multi-dataset Detection~\cite{zhou2021simple}, VinVL~\cite{Zhang_2021_CVPR}.}
\label{tab:od_result}
\end{table}

\begin{table*}[ht]
\centering
\setlength{\tabcolsep}{6.5pt}
\small
\renewcommand{\arraystretch}{1.35}
\begin{tabular}{cl|ccccccccccc}
\toprule
 & & \rotatebox{90}{Aquarium} & \rotatebox{90}{BCCD} & \rotatebox{90}{Chess Pieces} &
 \rotatebox{90}{Mask Wearing} & \rotatebox{90}{Oxford Pets} & \rotatebox{90}{Packages} &
 \rotatebox{90}{Pistols} & \rotatebox{90}{PKLot} & \rotatebox{90}{Pothole} & \rotatebox{90}{Thermal}
 & \rotatebox{90}{Wildfire Smoke} \\ \midrule
 &Images & 638 & 364 & 292 & 149 & 3680 & 26 & 2986 & 12416 & 665 & 203 & 737\\
 &Categories& 7 & 3 & 12 & 2 & 37 & 1 & 1 & 2 & 1 & 2 & 1\\
 \midrule\midrule
\multirow{2}{*}{\it{Fine-tuned}} & DyHead-Swin-L (full) & 53.1 & 62.6 & 80.7 &
52.0 & 85.9 & 52.0 & 74.4 & 98.0 & 61.8 & 75.9 & 58.7 \\
 & DyHead-Swin-L (5-shot) & 39.0 & 40.6 & 57.3 & 26.8 & 47.5 & 32.8 & 20.0 &
 22.1 & 10.8 & 54.9 & 14.2 \\  \midrule
\multirow{2}{*}{\it{Zero-shot}} & ZSD & 16.0 & 1.2 & 0.1 & 0.6 & 0.3 & 58.3
& 31.5 & 0.2 & 2.4 & 37.4 & 0.002 \\
 &Florence & 43.1 & 15.3 & 13.4 & 15.0 & 68.9 & 79.6 & 41.4 & 31.4 & 53.3 & 46.9 & 48.7 \\
 \bottomrule
\end{tabular}
\caption{Zero-shot transfer in object detection, in comparison with previous state-of-the-art
model DyHead~\cite{Dai_2021_CVPR} (on COCO) fine tuning results on full-set or 5-shot respectively and zero-shot detection baseline model ZSD~\cite{bansal2018zero}.}
\label{tab:zero_od_result}
\end{table*}
\begin{figure*}[ht]
  \centering
\includegraphics[width=1.0\linewidth]{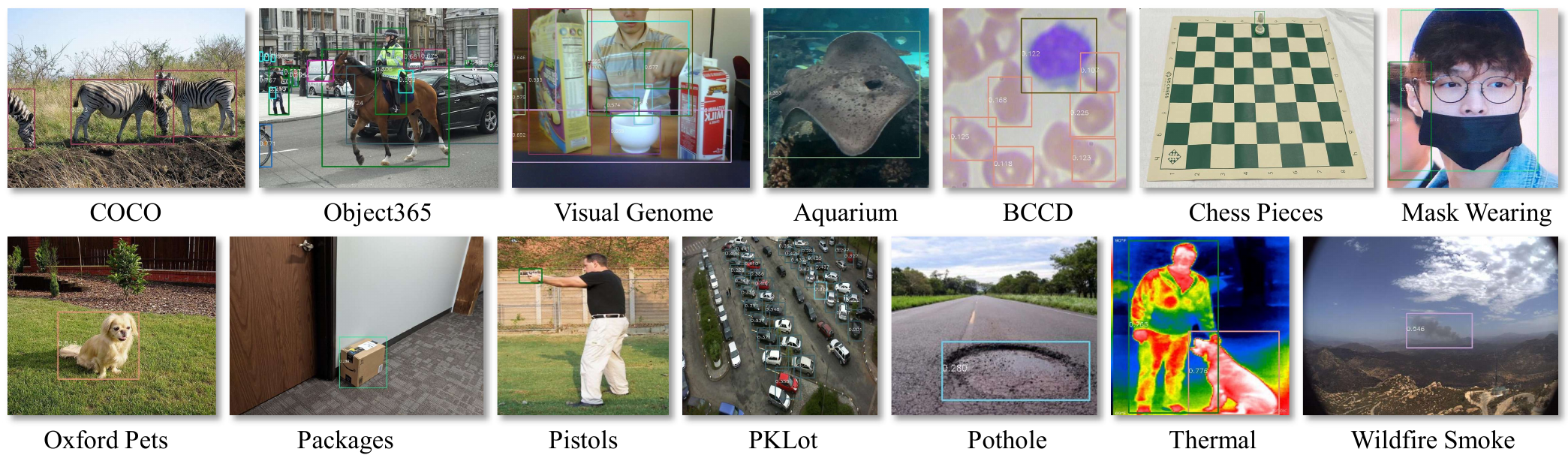} \vspace{-1.5em}
\caption{Our fine-tuned detection results on COCO (sparse object boxes), Object365 (dense object boxes), Visual Genome (w/ object attributes), and zero-shot transfer results on 11 downstream detection tasks. Boxes with different colors denote different object categories.}
\end{figure*}\vspace{0.3em}

We compare \emph{Florence} with state-of-the-art results on these three benchmarks in
Table~\ref{tab:od_result}. In object detection, the standard mean average precision (AP) metric is used to report results under different IoU thresholds and object scales for all datasets. We follow the metrics used in existing state-of-the-art methods. For COCO, Object365 and zero-shot transfer benchmarks, we use mAP, \ie, average over multiple IoUs ($0.5:0.05:0.95$). For Visual Genome, we use AP50 at IoU threshold $0.5$. As we can see, \emph{Florence} establishes new results in these main benchmarks of object detection.

\paragraph{Zero-shot Transfer} Zero-shot object detection is more challenging than zero-shot
classification, since neither object proposal classification nor location (\ie, bounding box
regression) in downstream tasks is seen during training. In our zero-shot transfer setting, object proposal and object classification are decoupled into two tasks. Object proposal discriminates object from
background, ignoring semantics of object categories. Classification, on the other hand, focuses on
object semantics for each bounding box proposal.
In spirit, this setup is similar to the behavior of R-CNN model~\cite{RCNN2014} which has been widely used for object detection before. Using this approach, we can follow existing work on zero-shot image classification to zero-shot
transfer in object detection, to evaluate the \emph{Florence} for novel object recognition. As mentioned in ZSD~\cite{bansal2018zero}, it more approaches real world settings.

For zero-shot transfer, the training of the detection adapter can be different from fine-tuning.
Specifically, we freeze the \emph{CoSwin} backbones and pre-train the \emph{Dynamic Head} on FLOD-9M
by neglecting semantics from each object bounding box. We treat the object detection pre-training as
general-purpose object proposal training. Note that the detection pre-training only updates the object adapter, and does not affect the fused
feature representations learned from large-scale image-text pairs. In inference, we apply the pre-trained \emph{CoSwin} and \emph{Dynamic Head} on
downstream datasets, and obtain the object proposals for every image. For each object proposal, we
apply zero-shot classification, as described in Section~\ref{sec:zero-shot}.

To evaluate \emph{Florence}'s transferability to novel, diverse and application-oriented tasks, following \cite{harold_GLIP2021}, we curate an ``open-set oject detection benchmark" which aggregates $11$  public datasets from
Roboflow\footnote{https://public.roboflow.com/object-detection}, spanning scenarios including fine-grained fishes/chess detection, drone-view detection, and thermal object detection. We use their split test datasets for evaluation. Table~\ref{tab:zero_od_result} shows that our \emph{Florence} model effectively zero-shot transfers to these tasks. We use the results of the baseline approach ZSD~\cite{bansal2018zero}, which considers a similar setting, for reference. In our implementation\footnote{We refer to \cite{harold_GLIP2021} for details.}, we replace their supervised object detector FasterRCNN with the recent SOTA detector~\cite{Dai_2021_CVPR} and use pre-trained BERT as the language encoder. Both are pre-trained end-to-end on the Objects365 dataset. Thanks to large-scale image-text pretraining, \emph{Florence} shows remarkable
gains on all tasks. Zero-shot in object detection still has a long way to be applied to
real-world tasks.
We further compare \emph{Florence} zero-shot with previous state-of-the-art
detector\footnote{It is pre-trained on ImageNet and COCO in supervised way.}~\cite{Dai_2021_CVPR} (on
COCO) fine-tunning on these tasks. We can observe noticeable performance gap between zero-shot and
supervised learning, especially for novel scenarios whose concepts/classes may not be covered by the pre-training dataset, such as ``BCCD" (blood cells photos), ``Chess Pieces" (Chess board photos and
various pieces). However, the results are  encouraging when compared with few-shot fine-tuning
results. \emph{Florence} outperforms in $7/11$ tasks over 5-shot fine tuning, and outperforms full-set fine-tuning on the ``Packages" dataset, consisting of only $26$ images for training. It demonstrates
the foundation models' great potential of improving data efficiency and reducing deployment cost for new tasks or domains.

\subsection{V+L Representation Learning}
The vision-langauge pretraining (VLP) is performed on
MSCOCO~\cite{lin2015microsoft}, Conceptual Captions (CC)~\cite{sharma2018conceptual},
CC12M~\cite{changpinyo2021cc12m},
SBU~\cite{OrdonezKB11}, and Visual Genome (VG)~\cite{krishnavisualgenome}.
These datasets result in $14$ million images with $20$ million associated captions.
Beyond replacing the image encoder with \emph{CoSwin}-H of
our \emph{Florence} model on~\cite{dou2021empirical}, we
remove the weight decay on the text embedding layer and
the modality-specific embedding.
ITM and MLM are applied for VLP with $43$ epochs with the image input size as $384$.

To evaluate the performance, we fine-tune the pre-trained model on the
challenging VQA~\cite{GoyalKSBP16} task, which is
to answer a question based on the image context.
The dataset consists of $82$K training images and $41$K validation images.
Only $1$K validation images are reserved and the rest are merged with
the training data for fine-tuning.
As a common practice, the problem is cast as a classification task
where each class corresponds to an answer.
The final pooling representations
are fed into a
randomly-initialized multilayer perceptron (MLP) network to predict the answer
over $3,129$ answers.
The loss is the binary cross-entropy loss, and the inference is to select the answer with the
highest confidence.
The model is fine-tuned for $10$ epochs with the learning rate as $8e-6$ and is evaluated
on the \texttt{test-dev} and \texttt{test-std}. The final accuracy is calculated on
the public server \footnote{http://evalai.com}.

Figure~\ref{tab:vqa} shows the comparison results with the existing methods.
As we can see, we achieve the new state-of-the-art performance.
Compared with SimVLM~\cite{wang2021simvlm}, which uses $1.8$B image-text pairs,
we only use $900$M data to pre-train the image encoder and $20$M
for VLP, but achieve better results.
This also demonstrates the data efficiency of our approach.

\begin{table}[t]
\centering
\setlength{\tabcolsep}{10.5pt}
\small
\renewcommand{\arraystretch}{1.35}
\begin{tabular}{l|cc}
\toprule
    Model                                 & test-dev            & test-std \\ \midrule
    UNITER~\cite{chen:uniter}           & 73.82               & 74.02 \\
    Visual Parsing~\cite{xue2021probing}  & 74.00               & 74.17 \\
    PixelBERT~\cite{huang2020pixel}       & 74.45               & 74.55 \\
    VILLA~\cite{gan:villa}             & 74.69               & 74.87 \\
    UNIMO~\cite{li2020unimo}              & 75.06               & 75.27 \\
    ALBEF~\cite{li2021align}              & 75.84               & 76.04 \\
    VinVL~\cite{Zhang_2021_CVPR}           & 76.52               & 76.60 \\
    CLIP-ViL~\cite{shen2021much}          & 76.48               & 76.70 \\
    METER~\cite{dou2021empirical}         & 77.68               & 77.64 \\
    SimVLM~\cite{wang2021simvlm}          & 80.03               & 80.34 \\
    \midrule
    Florence                            & \underline{\bf{80.16}}      & \underline{\bf{80.36}} \\
    \bottomrule
\end{tabular}
\caption{Compare our model with the existing state-of-the-art methods on VQA.}
  \label{tab:vqa}
\end{table}

\begin{table*}[ht]
\centering
\setlength{\tabcolsep}{10.5pt}
\small
\renewcommand{\arraystretch}{1.35}
\begin{tabular}{l|r|r|rrr}
\toprule
Method & Pre-training Type & Pre-training Data & R@1 & R@5 & R@10 \\
\midrule
MIL-NCE~\cite{miech2020end} & Video & HowTo100M & - & - & 32.4 \\
MMV~\cite{alayrac2020self} & Video & HowTo100M, AudioSet & - & - & 31.1 \\
VideoCLIP~\cite{xu2021videoclip} & Video$^*$ & HowTo100M & 10.4 & 22.2 & 30.0 \\
VATT~\cite{akbari2021vatt} & Video & HowTo100M, AudioSet & - & - & 29.7 \\
MCN~\cite{chen2021multimodal} & Image and Video & HowTo100M & - & - & 33.8 \\
Frozen-in-Time~\cite{bain2021frozen} & Image and Video & ImageNet,
CC, WebVid-2M & 18.7 & 39.5 & 51.6 \\
\midrule
CLIP-ViT-B/16~\cite{radford2021learning} & Image & WIT400M & 26.0 & 49.4
& 60.7 \\
Florence & Image & FLD-900M & \underline{\bf{37.6}} & \underline{\bf{63.8}} & \underline{\bf{72.6}}
\\
\bottomrule
\end{tabular}
\caption{Zero-shot text-to-video retrieval results on MSR-VTT 1K-A test set. ($^*$: Feature extracted from the pre-trained model~\cite{miech2020end}, followed by another stage of video-and-language pre-training) The pretraining data used in these existing methods include HowTo100M~\cite{miech2019howto100m}, AudioSet~\cite{gemmeke2017audio},  ImageNet~\cite{deng2009imagenet}, CC~\cite{sharma2018conceptual},  WebVid-2M~\cite{bain2021frozen}, WIT400M~\cite{radford2021learning}}
\label{tab:t2v_retrieval_results}
\end{table*}

\begin{table*}[ht]
\centering
\setlength{\tabcolsep}{12pt}
\small
\renewcommand{\arraystretch}{1.35}
\begin{tabular}{l|r|cc|cc|c|c}
\toprule
\multirow{2}{*}{Method} & \multirow{2}{*}{Pretraining Data} & \multicolumn{2}{c|}{Kinetics-400} &
\multicolumn{2}{c|}{Kinetics-600} & \multirow{2}{*}{Views} & \multirow{2}{*}{Params}\\
& & Top-1 & Top-5 & Top-1 & Top-5 & & \\
\midrule
ViViT-H/16x2 & JFT-300M & 84.8 & 95.8 & 85.8 & 96.5 & 4 $\times$ 3 & 648M\\
VideoSwin-L & ImageNet-22K & 84.6 & 96.5 & 85.9 & 97.1 & 4 $\times$ 3 & 200M \\
VideoSwin-L & ImageNet-22K & 84.9 & 96.7 & 86.1 & 97.3 & 10 $\times$ 5 & 200M
\\
TokenLearner 16at18+L/10 & JFT-300M & 85.4 & 96.3 & 86.3 & 97.0 & 4
$\times$ 3 & 460M \\
Florence & FLD-900M & \underline{\bf{86.5}} & \underline{\bf{97.3}} & \underline{\bf{87.8}} &
\underline{\bf{97.8}} & 4 $\times$ 3 & 647M \\
\bottomrule
\end{tabular}
\caption{Comparison to state-of-the-art methods, including ViViT~\cite{arnab2021vivit}, VideoSwin~\cite{liu2021video}, TokenLearner~\cite{ryoo2021tokenlearner}, on Kinetics-400 and Kinetics-600. Views indicate $\#
temporal\;clip \times \# spatial\;crop$. }
\label{tab:k400}
\end{table*}

\subsection{Zero-Shot Text-to-Video Retrieval}

Although \emph{Florence} is pre-trained on image-text pairs, it can be easily adapted to video tasks
(shown in Section~\ref{sect:adapt_video}), such as text-video retrieval. We expand the input 2D
patch embeddings and positional embeddings to 3D so that the encoder can process video inputs,
following~\cite{arnab2021vivit}. Then, we perform zero-shot text-to-video evaluation on the
MSR-VTT~\cite{xu2016msr} dataset. We report results on the 1K-A test~\cite{yu2018joint}, which
contains 1K video and caption pairs. We use the standard recall metrics for evaluation and compare
with existing state-of-the-art methods in Table~\ref{tab:t2v_retrieval_results}. As we can see, these
two image-text pre-trained models CLIP\footnote{We use a public available CLIP checkpoint for
comparison}~\cite{radford2021learning} and \emph{Florence} outperform all the state-of-the-art
methods by a large margin in terms of the $R@1$ metric. It reveals that the video data used for
pretraining in these state-of-the-art methods may not be so rich or diverse as image-text data used
in \emph{Florence} or CLIP.

\subsection{Video Action Recognition}

We evaluate \emph{Florence} on fine-tuned video action recognition tasks. On the Kinectics-400 and
Kinectics-600 datasets, we follow the typical fine-tuning setting~\cite{liu2021video} and fine tune
the model (Section~\ref{sect:adapt_video}) with $384\times 384$ resolution for $30$ epochs. We use
the label smoothing, rand augmentation, a small learning rate $0.0002$ and a relatively large drop
path rate $0.5$ to avoid over-fitting the target video datasets. We compare with existing
state-of-the-art methods in Table~\ref{tab:k400}. Our results are better than the state-of-the-art
by $1.1\%$ and $1.5\%$ on Kinectics-400 and Kinectics-600, respectively.


\section{Conclusion and Future Work}

In this paper we investigated a new paradigm of building a computer vision foundation model, \emph{Florence}, as a general-purpose vision system. Our attempt is a step towards building XYZ-code \cite{XYZ-code}, an integrative AI system that makes progress toward human-like AI. Although the model size is still below several other existing billion-scale models, \emph{Florence} successfully extends to different tasks along space, time, and modality, with great transferbility, and achieves new SOTA results on a wide range of vision benchmarks.

For the future work, we plan to include more vision tasks and applications, such as depth/flow estimation, tracking, and additional vision+language tasks. \emph{Florence} is designed to pave the way for building vision foundation models to power millions of real-world vision tasks and applications. In addition, the preliminary progress on zero-shot classification and object detection may motivate more research to close the performance gap to supervised learning.

\noindent\paragraph{\textsc{Acknowledgment}}\mbox{}\\

We would like to thank the following people involved in the discussion for their valuable feedback including Xiaowei Hu, Yen-Chun Chen, Lin Liang, Yinpeng Chen, Li Dong, Furu Wei, Han Hu, Yue Cao, Zheng Zhang, Hao Yang, Jianmin Bao, Dong Chen, Fang Wen, Jianlong Fu, Houwen Peng, Chong Luo, Baining Guo. We would also thank Qingfen Lin, Cha Zhang for their thoughtful feedback on the broader impacts of the paper. Thanks Mei Gao, Ping Jin for helping run evaluations on benchmark infrastructure. We are also grateful to the developers
of software toolkits used throughout this project, including Liyang Lu, Robert Gmyr, Felipe Cruz Salinas, Canrun Li, Steven Tsai, Min Gao, Kevin Pan, Shohei Ono, Christina Sun. Additionally, we would like to thank the entire Deepspeed, AI Frameworks, and ITP teams for making it possible to train models at this scale.

\bibliography{egbib}

\begin{thebibliography}{88}
\providecommand{\natexlab}[1]{#1}
\providecommand{\url}[1]{\texttt{#1}}
\expandafter\ifx\csname urlstyle\endcsname\relax
  \providecommand{\doi}[1]{doi: #1}\else
  \providecommand{\doi}{doi: \begingroup \urlstyle{rm}\Url}\fi

\bibitem[Wud()]{Wudao2}
Wu dao 2.0.
\newblock \url{https://gpt3demo.com/apps/wu-dao-20}.

\bibitem[Akbari et~al.(2021)Akbari, Yuan, Qian, Chuang, Chang, Cui, and
  Gong]{akbari2021vatt}
Akbari, H., Yuan, L., Qian, R., Chuang, W.-H., Chang, S.-F., Cui, Y., and Gong,
  B.
\newblock Vatt: Transformers for multimodal self-supervised learning from raw
  video, audio and text.
\newblock In \emph{NeurIPS}, 2021.

\bibitem[Alayrac et~al.(2020)Alayrac, Recasens, Schneider, Arandjelovic,
  Ramapuram, De~Fauw, Smaira, Dieleman, and Zisserman]{alayrac2020self}
Alayrac, J.-B., Recasens, A., Schneider, R., Arandjelovic, R., Ramapuram, J.,
  De~Fauw, J., Smaira, L., Dieleman, S., and Zisserman, A.
\newblock Self-supervised multimodal versatile networks.
\newblock In \emph{NeurIPS}, volume~2, pp.\ ~7, 2020.

\bibitem[Anderson et~al.(2018)Anderson, He, Buehler, Teney, Johnson, Gould, and
  Zhang]{00010BT0GZ18}
Anderson, P., He, X., Buehler, C., Teney, D., Johnson, M., Gould, S., and
  Zhang, L.
\newblock Bottom-up and top-down attention for image captioning and visual
  question answering.
\newblock In \emph{CVPR}, 2018.

\bibitem[Arnab et~al.(2021)Arnab, Dehghani, Heigold, Sun, Lu{\v{c}}i{\'c}, and
  Schmid]{arnab2021vivit}
Arnab, A., Dehghani, M., Heigold, G., Sun, C., Lu{\v{c}}i{\'c}, M., and Schmid,
  C.
\newblock Vivit: A video vision transformer.
\newblock In \emph{ICCV}, 2021.

\bibitem[Bain et~al.(2021)Bain, Nagrani, Varol, and Zisserman]{bain2021frozen}
Bain, M., Nagrani, A., Varol, G., and Zisserman, A.
\newblock Frozen in time: A joint video and image encoder for end-to-end
  retrieval.
\newblock In \emph{ICCV}, 2021.

\bibitem[Bansal et~al.(2018)Bansal, Sikka, Sharma, Chellappa, and
  Divakaran]{bansal2018zero}
Bansal, A., Sikka, K., Sharma, G., Chellappa, R., and Divakaran, A.
\newblock Zero-shot object detection.
\newblock In \emph{Proceedings of the European Conference on Computer Vision
  (ECCV)}, pp.\  384--400, 2018.

\bibitem[Berg et~al.(2014)Berg, Liu, Lee, Alexander, Jacobs, and
  Belhumeur]{6909656}
Berg, T., Liu, J., Lee, S.~W., Alexander, M.~L., Jacobs, D.~W., and Belhumeur,
  P.~N.
\newblock Birdsnap: Large-scale fine-grained visual categorization of birds.
\newblock In \emph{2014 IEEE Conference on Computer Vision and Pattern
  Recognition}, pp.\  2019--2026, 2014.

\bibitem[Bommasani et~al.(2021)Bommasani, Hudson, Adeli, Altman, Arora, von
  Arx, Bernstein, Bohg, Bosselut, Brunskill, Brynjolfsson, Buch, Card,
  Castellon, Chatterji, Chen, Creel, Davis, Demszky, Donahue, Doumbouya,
  Durmus, Ermon, Etchemendy, Ethayarajh, Fei-Fei, Finn, Gale, Gillespie, Goel,
  Goodman, Grossman, Guha, Hashimoto, Henderson, Hewitt, Ho, Hong, Hsu, Huang,
  Icard, Jain, Jurafsky, Kalluri, Karamcheti, Keeling, Khani, Khattab, Koh,
  Krass, Krishna, Kuditipudi, Kumar, Ladhak, Lee, Lee, Leskovec, Levent, Li,
  Li, Ma, Malik, Manning, Mirchandani, Mitchell, Munyikwa, Nair, Narayan,
  Narayanan, Newman, Nie, Niebles, Nilforoshan, Nyarko, Ogut, Orr,
  Papadimitriou, Park, Piech, Portelance, Potts, Raghunathan, Reich, Ren, Rong,
  Roohani, Ruiz, Ryan, Ré, Sadigh, Sagawa, Santhanam, Shih, Srinivasan,
  Tamkin, Taori, Thomas, Tramèr, Wang, Wang, Wu, Wu, Wu, Xie, Yasunaga, You,
  Zaharia, Zhang, Zhang, Zhang, Zhang, Zheng, Zhou, and
  Liang]{bommasani2021opportunities}
Bommasani, R., Hudson, D.~A., Adeli, E., Altman, R., Arora, S., von Arx, S.,
  Bernstein, M.~S., Bohg, J., Bosselut, A., Brunskill, E., Brynjolfsson, E.,
  Buch, S., Card, D., Castellon, R., Chatterji, N., Chen, A., Creel, K., Davis,
  J.~Q., Demszky, D., Donahue, C., Doumbouya, M., Durmus, E., Ermon, S.,
  Etchemendy, J., Ethayarajh, K., Fei-Fei, L., Finn, C., Gale, T., Gillespie,
  L., Goel, K., Goodman, N., Grossman, S., Guha, N., Hashimoto, T., Henderson,
  P., Hewitt, J., Ho, D.~E., Hong, J., Hsu, K., Huang, J., Icard, T., Jain, S.,
  Jurafsky, D., Kalluri, P., Karamcheti, S., Keeling, G., Khani, F., Khattab,
  O., Koh, P.~W., Krass, M., Krishna, R., Kuditipudi, R., Kumar, A., Ladhak,
  F., Lee, M., Lee, T., Leskovec, J., Levent, I., Li, X.~L., Li, X., Ma, T.,
  Malik, A., Manning, C.~D., Mirchandani, S., Mitchell, E., Munyikwa, Z., Nair,
  S., Narayan, A., Narayanan, D., Newman, B., Nie, A., Niebles, J.~C.,
  Nilforoshan, H., Nyarko, J., Ogut, G., Orr, L., Papadimitriou, I., Park,
  J.~S., Piech, C., Portelance, E., Potts, C., Raghunathan, A., Reich, R., Ren,
  H., Rong, F., Roohani, Y., Ruiz, C., Ryan, J., Ré, C., Sadigh, D., Sagawa,
  S., Santhanam, K., Shih, A., Srinivasan, K., Tamkin, A., Taori, R., Thomas,
  A.~W., Tramèr, F., Wang, R.~E., Wang, W., Wu, B., Wu, J., Wu, Y., Xie,
  S.~M., Yasunaga, M., You, J., Zaharia, M., Zhang, M., Zhang, T., Zhang, X.,
  Zhang, Y., Zheng, L., Zhou, K., and Liang, P.
\newblock On the opportunities and risks of foundation models.
\newblock In \emph{arXiv 2108.07258}, 2021.

\bibitem[Brown et~al.(2020)Brown, Mann, Ryder, Subbiah, Kaplan, Dhariwal,
  Neelakantan, Shyam, Sastry, Askell, Agarwal, Herbert-Voss, Krueger, Henighan,
  Child, Ramesh, Ziegler, Wu, Winter, Hesse, Chen, Sigler, Litwin, Gray, Chess,
  Clark, Berner, McCandlish, Radford, Sutskever, and Amodei]{brown2020language}
Brown, T.~B., Mann, B., Ryder, N., Subbiah, M., Kaplan, J., Dhariwal, P.,
  Neelakantan, A., Shyam, P., Sastry, G., Askell, A., Agarwal, S.,
  Herbert-Voss, A., Krueger, G., Henighan, T., Child, R., Ramesh, A., Ziegler,
  D.~M., Wu, J., Winter, C., Hesse, C., Chen, M., Sigler, E., Litwin, M., Gray,
  S., Chess, B., Clark, J., Berner, C., McCandlish, S., Radford, A., Sutskever,
  I., and Amodei, D.
\newblock Language models are few-shot learners.
\newblock In \emph{arXiv 2005.14165}, 2020.

\bibitem[Changpinyo et~al.(2021)Changpinyo, Sharma, Ding, and
  Soricut]{changpinyo2021cc12m}
Changpinyo, S., Sharma, P., Ding, N., and Soricut, R.
\newblock {Conceptual 12M}: Pushing web-scale image-text pre-training to
  recognize long-tail visual concepts.
\newblock In \emph{CVPR}, 2021.

\bibitem[Chen et~al.(2021)Chen, Rouditchenko, Duarte, Kuehne, Thomas, Boggust,
  Panda, Kingsbury, Feris, Harwath, et~al.]{chen2021multimodal}
Chen, B., Rouditchenko, A., Duarte, K., Kuehne, H., Thomas, S., Boggust, A.,
  Panda, R., Kingsbury, B., Feris, R., Harwath, D., et~al.
\newblock Multimodal clustering networks for self-supervised learning from
  unlabeled videos.
\newblock In \emph{ICCV}, 2021.

\bibitem[Chen et~al.(2020{\natexlab{a}})Chen, Hu, Wu, Jiang, and
  Wang]{chen:vsepooling}
Chen, J., Hu, H., Wu, H., Jiang, Y., and Wang, C.
\newblock Learning the best pooling strategy for visual semantic embedding.
\newblock In \emph{arXiv preprint arXiv:2011.04305}, 2020{\natexlab{a}}.

\bibitem[Chen et~al.(2020{\natexlab{b}})Chen, Kornblith, Norouzi, and
  Hinton]{pmlr-v119-chen20j}
Chen, T., Kornblith, S., Norouzi, M., and Hinton, G.
\newblock A simple framework for contrastive learning of visual
  representations.
\newblock In \emph{Proceedings of the 37th International Conference on Machine
  Learning}, volume 119, pp.\  1597--1607, 13--18 Jul 2020{\natexlab{b}}.

\bibitem[Chen et~al.(2020{\natexlab{c}})Chen, Kornblith, Swersky, Norouzi, and
  Hinton]{chen2020big}
Chen, T., Kornblith, S., Swersky, K., Norouzi, M., and Hinton, G.
\newblock Big self-supervised models are strong semi-supervised learners.
\newblock \emph{arXiv preprint arXiv:2006.10029}, 2020{\natexlab{c}}.

\bibitem[Chen et~al.(2020{\natexlab{d}})Chen, Li, Yu, Kholy, Ahmed, Gan, Cheng,
  and Liu]{chen:uniter}
Chen, Y.-C., Li, L., Yu, L., Kholy, A.~E., Ahmed, F., Gan, Z., Cheng, Y., and
  Liu, J.
\newblock Uniter: Universal image-text representation learning.
\newblock In \emph{Proceedings of European Conference on Computer Vision},
  2020{\natexlab{d}}.

\bibitem[Codella et~al.(2019)Codella, Rotemberg, Tschandl, Celebi, Dusza,
  Gutman, Helba, Kalloo, Liopyris, Marchetti, Kittler, and Halpern]{isic2018}
Codella, N. C.~F., Rotemberg, V., Tschandl, P., Celebi, M.~E., Dusza, S.~W.,
  Gutman, D.~A., Helba, B., Kalloo, A., Liopyris, K., Marchetti, M.~A.,
  Kittler, H., and Halpern, A.
\newblock Skin lesion analysis toward melanoma detection 2018: {A} challenge
  hosted by the international skin imaging collaboration {(ISIC)}.
\newblock abs/1902.03368, 2019.

\bibitem[Dai et~al.(2021{\natexlab{a}})Dai, Chen, Xiao, Chen, Liu, Yuan, and
  Zhang]{Dai_2021_CVPR}
Dai, X., Chen, Y., Xiao, B., Chen, D., Liu, M., Yuan, L., and Zhang, L.
\newblock Dynamic head: Unifying object detection heads with attentions.
\newblock In \emph{Proceedings of the IEEE/CVF Conference on Computer Vision
  and Pattern Recognition (CVPR)}, pp.\  7373--7382, June 2021{\natexlab{a}}.

\bibitem[Dai et~al.(2021{\natexlab{b}})Dai, Chen, Yang, Zhang, Yuan, and
  Zhang]{Dai_2021_ICCV}
Dai, X., Chen, Y., Yang, J., Zhang, P., Yuan, L., and Zhang, L.
\newblock Dynamic detr: End-to-end object detection with dynamic attention.
\newblock In \emph{Proceedings of the IEEE/CVF International Conference on
  Computer Vision (ICCV)}, pp.\  2988--2997, October 2021{\natexlab{b}}.

\bibitem[Dai et~al.(2021{\natexlab{c}})Dai, Liu, Le, and Tan]{dai2021coatnet}
Dai, Z., Liu, H., Le, Q.~V., and Tan, M.
\newblock Coatnet: Marrying convolution and attention for all data sizes.
\newblock In \emph{arXiv 2106.04803}, 2021{\natexlab{c}}.

\bibitem[Dean()]{pathways}
Dean, J.
\newblock Introducing pathways: A next-generation ai architecture.
\newblock
  \url{https://blog.google/technology/ai/introducing-pathways-next-generation-ai-architecture/}.

\bibitem[Deng et~al.(2009)Deng, Dong, Socher, Li, Li, and
  Fei-Fei]{deng2009imagenet}
Deng, J., Dong, W., Socher, R., Li, L.-J., Li, K., and Fei-Fei, L.
\newblock Imagenet: A large-scale hierarchical image database.
\newblock In \emph{2009 IEEE conference on computer vision and pattern
  recognition}, pp.\  248--255. Ieee, 2009.

\bibitem[Devlin et~al.(2019)Devlin, Chang, Lee, and Toutanova]{devlin2019bert}
Devlin, J., Chang, M.-W., Lee, K., and Toutanova, K.
\newblock Bert: Pre-training of deep bidirectional transformers for language
  understanding.
\newblock In \emph{arXiv 1810.04805}, 2019.

\bibitem[Dong et~al.(2021)Dong, Bao, Chen, Zhang, Yu, Yuan, Chen, and
  Guo]{dong2021cswin}
Dong, X., Bao, J., Chen, D., Zhang, W., Yu, N., Yuan, L., Chen, D., and Guo, B.
\newblock Cswin transformer: A general vision transformer backbone with
  cross-shaped windows.
\newblock In \emph{arXiv 2107.00652}, 2021.

\bibitem[Dosovitskiy et~al.(2021{\natexlab{a}})Dosovitskiy, Beyer, Kolesnikov,
  Weissenborn, Zhai, Unterthiner, Dehghani, Minderer, Heigold, Gelly,
  Uszkoreit, and Houlsby]{dosovitskiy2020vit}
Dosovitskiy, A., Beyer, L., Kolesnikov, A., Weissenborn, D., Zhai, X.,
  Unterthiner, T., Dehghani, M., Minderer, M., Heigold, G., Gelly, S.,
  Uszkoreit, J., and Houlsby, N.
\newblock An image is worth 16x16 words: Transformers for image recognition at
  scale.
\newblock \emph{ICLR}, 2021{\natexlab{a}}.

\bibitem[Dosovitskiy et~al.(2021{\natexlab{b}})Dosovitskiy, Beyer, Kolesnikov,
  Weissenborn, Zhai, Unterthiner, Dehghani, Minderer, Heigold, Gelly,
  Uszkoreit, and Houlsby]{dosovitskiy2021image}
Dosovitskiy, A., Beyer, L., Kolesnikov, A., Weissenborn, D., Zhai, X.,
  Unterthiner, T., Dehghani, M., Minderer, M., Heigold, G., Gelly, S.,
  Uszkoreit, J., and Houlsby, N.
\newblock An image is worth 16x16 words: Transformers for image recognition at
  scale.
\newblock In \emph{arXiv 2010.11929}, 2021{\natexlab{b}}.

\bibitem[Dou et~al.(2021)Dou, Xu, Gan, Wang, Wang, Wang, Zhu, Nanyun, Peng,
  Liu, and Zeng]{dou2021empirical}
Dou, Z.-Y., Xu, Y., Gan, Z., Wang, J., Wang, S., Wang, L., Zhu, C., Nanyun,
  Peng, Liu, Z., and Zeng, M.
\newblock An empirical study of training end-to-end vision-and-language
  transformers.
\newblock In \emph{arXiv 2111.02387}, 2021.

\bibitem[Fang et~al.(2021)Fang, Wang, Hu, Wang, Yang, and Liu]{abs-2104-02096}
Fang, Z., Wang, J., Hu, X., Wang, L., Yang, Y., and Liu, Z.
\newblock Compressing visual-linguistic model via knowledge distillation.
\newblock In \emph{ICCV}, 2021.

\bibitem[Gan et~al.(2020)Gan, Chen, Li, Zhu, Cheng, and Liu]{gan:villa}
Gan, Z., Chen, Y.-C., Li, L., Zhu, C., Cheng, Y., and Liu, J.
\newblock Large-scale adversarial training for vision-and-language
  representation learning.
\newblock In \emph{Proceedings of Neural Information Processing Systems}, 2020.

\bibitem[Gao et~al.(2021)Gao, Zhang, Han, and Callan]{gao2021scaling}
Gao, L., Zhang, Y., Han, J., and Callan, J.
\newblock Scaling deep contrastive learning batch size under memory limited
  setup.
\newblock In \emph{arXiv 2101.06983}, 2021.

\bibitem[Gemmeke et~al.(2017)Gemmeke, Ellis, Freedman, Jansen, Lawrence, Moore,
  Plakal, and Ritter]{gemmeke2017audio}
Gemmeke, J.~F., Ellis, D.~P., Freedman, D., Jansen, A., Lawrence, W., Moore,
  R.~C., Plakal, M., and Ritter, M.
\newblock Audio set: An ontology and human-labeled dataset for audio events.
\newblock In \emph{ICASSP}, pp.\  776--780. IEEE, 2017.

\bibitem[Girshick et~al.(2014)Girshick, Donahue, Darrell, and Malik]{RCNN2014}
Girshick, R., Donahue, J., Darrell, T., and Malik, J.
\newblock Rich feature hierarchies for accurate object detection and semantic
  segmentation.
\newblock In \emph{2014 IEEE Conference on Computer Vision and Pattern
  Recognition}, pp.\  580--587, 2014.

\bibitem[Goyal et~al.(2017)Goyal, Khot, Summers{-}Stay, Batra, and
  Parikh]{GoyalKSBP16}
Goyal, Y., Khot, T., Summers{-}Stay, D., Batra, D., and Parikh, D.
\newblock Making the {V} in {VQA} matter: Elevating the role of image
  understanding in visual question answering.
\newblock In \emph{CVPR}, 2017.

\bibitem[Guo et~al.(2020)Guo, Codella, Karlinsky, Smith, Rosing, and
  Feris]{cdfsl}
Guo, Y., Codella, N. C.~F., Karlinsky, L., Smith, J.~R., Rosing, T., and Feris,
  R.~S.
\newblock A new benchmark for evaluation of cross-domain few-shot learning.
\newblock \emph{ECCV}, 2020.

\bibitem[Gupta et~al.(2019)Gupta, Dollar, and Girshick]{Gupta_2019_CVPR}
Gupta, A., Dollar, P., and Girshick, R.
\newblock Lvis: A dataset for large vocabulary instance segmentation.
\newblock In \emph{Proceedings of the IEEE/CVF Conference on Computer Vision
  and Pattern Recognition (CVPR)}, June 2019.

\bibitem[Helber et~al.(2019)Helber, Bischke, Dengel, and Borth]{eurosat}
Helber, P., Bischke, B., Dengel, A., and Borth, D.
\newblock Eurosat: A novel dataset and deep learning benchmark for land use and
  land cover classification.
\newblock \emph{IEEE Journal of Selected Topics in Applied Earth Observations
  and Remote Sensing}, 12\penalty0 (7):\penalty0 2217--2226, 2019.

\bibitem[Huang()]{XYZ-code}
Huang, X.
\newblock A holistic representation toward integrative ai.
\newblock
  \url{https://www.microsoft.com/en-us/research/blog/a-holistic-representation-toward-integrative-ai/}.

\bibitem[Huang et~al.(2020)Huang, Zeng, Liu, Fu, and Fu]{huang2020pixel}
Huang, Z., Zeng, Z., Liu, B., Fu, D., and Fu, J.
\newblock Pixel-{BERT}: Aligning image pixels with text by deep multi-modal
  transformers.
\newblock \emph{arXiv preprint}, 2020.

\bibitem[Huang et~al.(2021)Huang, Zeng, Huang, Liu, Fu, and Fu]{abs-2104-03135}
Huang, Z., Zeng, Z., Huang, Y., Liu, B., Fu, D., and Fu, J.
\newblock Seeing out of the box: End-to-end pre-training for vision-language
  representation learning.
\newblock In \emph{CVPR}, 2021.

\bibitem[Jia et~al.(2021)Jia, Yang, Xia, Chen, Parekh, Pham, Le, Sung, Li, and
  Duerig]{jia2021scaling}
Jia, C., Yang, Y., Xia, Y., Chen, Y.-T., Parekh, Z., Pham, H., Le, Q.~V., Sung,
  Y., Li, Z., and Duerig, T.
\newblock Scaling up visual and vision-language representation learning with
  noisy text supervision.
\newblock In \emph{arXiv 2102.05918}, 2021.

\bibitem[Kim et~al.(2021)Kim, Son, and Kim]{KimSK21}
Kim, W., Son, B., and Kim, I.
\newblock Vilt: Vision-and-language transformer without convolution or region
  supervision.
\newblock In Meila, M. and Zhang, T. (eds.), \emph{ICML}, 2021.

\bibitem[Kolesnikov et~al.(2020)Kolesnikov, Beyer, Zhai, Puigcerver, Yung,
  Gelly, and Houlsby]{kolesnikov2020big}
Kolesnikov, A., Beyer, L., Zhai, X., Puigcerver, J., Yung, J., Gelly, S., and
  Houlsby, N.
\newblock Big transfer (bit): General visual representation learning.
\newblock In \emph{arXiv 1912.11370}, 2020.

\bibitem[Kornblith et~al.(2019)Kornblith, Shlens, and Le]{kornblith2019better}
Kornblith, S., Shlens, J., and Le, Q.~V.
\newblock Do better imagenet models transfer better?
\newblock In \emph{Proceedings of the IEEE/CVF Conference on Computer Vision
  and Pattern Recognition}, pp.\  2661--2671, 2019.

\bibitem[Krasin et~al.(2016)Krasin, Duerig, Alldrin, Veit, Abu-El-Haija,
  Belongie, Cai, Feng, Ferrari, Gomes, Gupta, Narayanan, Sun, Chechik, and
  Murphy]{openimages}
Krasin, I., Duerig, T., Alldrin, N., Veit, A., Abu-El-Haija, S., Belongie, S.,
  Cai, D., Feng, Z., Ferrari, V., Gomes, V., Gupta, A., Narayanan, D., Sun, C.,
  Chechik, G., and Murphy, K.
\newblock Openimages: A public dataset for large-scale multi-label and
  multi-class image classification.
\newblock \emph{Dataset available from https://github.com/openimages}, 2016.

\bibitem[Krishna et~al.(2016)Krishna, Zhu, Groth, Johnson, Hata, Kravitz, Chen,
  Kalantidis, Li, Shamma, Bernstein, and Fei-Fei]{krishnavisualgenome}
Krishna, R., Zhu, Y., Groth, O., Johnson, J., Hata, K., Kravitz, J., Chen, S.,
  Kalantidis, Y., Li, L.-J., Shamma, D.~A., Bernstein, M., and Fei-Fei, L.
\newblock Visual genome: Connecting language and vision using crowdsourced
  dense image annotations.
\newblock In \emph{arXiv 1602.07332}, 2016.

\bibitem[Li et~al.(2021{\natexlab{a}})Li, Selvaraju, Gotmare, Joty, Xiong, and
  Hoi]{li2021align}
Li, J., Selvaraju, R.~R., Gotmare, A.~D., Joty, S., Xiong, C., and Hoi, S.
\newblock Align before fuse: Vision and language representation learning with
  momentum distillation.
\newblock In \emph{Conference on Neural Information Processing Systems
  (NeurIPS)}, 2021{\natexlab{a}}.

\bibitem[Li et~al.(2021{\natexlab{b}})Li, Zhang, Zhang, Yang, Li, Zhong, Wang,
  Yuan, Zhang, Hwang, Chang, and Gao]{harold_GLIP2021}
Li, L.~H., Zhang, P., Zhang, H., Yang, J., Li, C., Zhong, Y., Wang, L., Yuan,
  L., Zhang, L., Hwang, J.-N., Chang, K.-W., and Gao, J.
\newblock Grounded language-image pre-training.
\newblock In \emph{arXiv In Preparation}, 2021{\natexlab{b}}.

\bibitem[Li et~al.(2021{\natexlab{c}})Li, Gao, Niu, Xiao, Liu, Liu, Wu, and
  Wang]{li2020unimo}
Li, W., Gao, C., Niu, G., Xiao, X., Liu, H., Liu, J., Wu, H., and Wang, H.
\newblock Unimo: Towards unified-modal understanding and generation via
  cross-modal contrastive learning.
\newblock In \emph{Annual Meeting of the Association for Computational
  Linguistics (ACL)}, 2021{\natexlab{c}}.

\bibitem[Li et~al.(2020)Li, Yin, Li, Zhang, Hu, Zhang, Wang, Hu, Dong, Wei,
  Choi, and Gao]{li:oscar}
Li, X., Yin, X., Li, C., Zhang, P., Hu, X., Zhang, L., Wang, L., Hu, H., Dong,
  L., Wei, F., Choi, Y., and Gao, J.
\newblock Oscar: Object-semantics aligned pre-training for vision-language
  tasks.
\newblock In \emph{Proceedings of European Conference on Computer Vision},
  2020.

\bibitem[Lin et~al.(2015)Lin, Maire, Belongie, Bourdev, Girshick, Hays, Perona,
  Ramanan, Zitnick, and Dollár]{lin2015microsoft}
Lin, T.-Y., Maire, M., Belongie, S., Bourdev, L., Girshick, R., Hays, J.,
  Perona, P., Ramanan, D., Zitnick, C.~L., and Dollár, P.
\newblock Microsoft {COCO:}: Common objects in context, 2015.

\bibitem[Liu et~al.(2020)Liu, Zhao, Li, Jiang, Guo, and Ye]{cdfsltop}
Liu, B., Zhao, Z., Li, Z., Jiang, J., Guo, Y., and Ye, J.
\newblock Feature transformation ensemble model with batch spectral
  regularization for cross-domain few-shot classification.
\newblock 2020.

\bibitem[Liu et~al.(2019)Liu, Ott, Goyal, Du, Joshi, Chen, Levy, Lewis,
  Zettlemoyer, and Stoyanov]{liu2019roberta}
Liu, Y., Ott, M., Goyal, N., Du, J., Joshi, M., Chen, D., Levy, O., Lewis, M.,
  Zettlemoyer, L., and Stoyanov, V.
\newblock Ro{BERT}a: A robustly optimized bert pretraining approach.
\newblock \emph{arXiv preprint}, 2019.

\bibitem[Liu et~al.(2021{\natexlab{a}})Liu, Lin, Cao, Hu, Wei, Zhang, Lin, and
  Guo]{liu2021Swin}
Liu, Z., Lin, Y., Cao, Y., Hu, H., Wei, Y., Zhang, Z., Lin, S., and Guo, B.
\newblock Swin transformer: Hierarchical vision transformer using shifted
  windows.
\newblock \emph{International Conference on Computer Vision (ICCV)},
  2021{\natexlab{a}}.

\bibitem[Liu et~al.(2021{\natexlab{b}})Liu, Ning, Cao, Wei, Zhang, Lin, and
  Hu]{liu2021video}
Liu, Z., Ning, J., Cao, Y., Wei, Y., Zhang, Z., Lin, S., and Hu, H.
\newblock Video swin transformer.
\newblock \emph{arXiv preprint arXiv:2106.13230}, 2021{\natexlab{b}}.

\bibitem[Miech et~al.(2019)Miech, Zhukov, Alayrac, Tapaswi, Laptev, and
  Sivic]{miech2019howto100m}
Miech, A., Zhukov, D., Alayrac, J.-B., Tapaswi, M., Laptev, I., and Sivic, J.
\newblock Howto100m: Learning a text-video embedding by watching hundred
  million narrated video clips.
\newblock In \emph{ICCV}, pp.\  2630--2640, 2019.

\bibitem[Miech et~al.(2020)Miech, Alayrac, Smaira, Laptev, Sivic, and
  Zisserman]{miech2020end}
Miech, A., Alayrac, J.-B., Smaira, L., Laptev, I., Sivic, J., and Zisserman, A.
\newblock End-to-end learning of visual representations from uncurated
  instructional videos.
\newblock In \emph{CVPR}, pp.\  9879--9889, 2020.

\bibitem[Mohanty et~al.(2016)Mohanty, Hughes, and Salathe]{plantdisease}
Mohanty, S.~P., Hughes, D.~P., and Salathe, M.
\newblock Using deep learning for image-based plant disease detection.
\newblock \emph{Front Plant Sci}, 7, 2016.

\bibitem[Ordonez et~al.(2011)Ordonez, Kulkarni, and Berg]{OrdonezKB11}
Ordonez, V., Kulkarni, G., and Berg, T.~L.
\newblock Im2text: Describing images using 1 million captioned photographs.
\newblock In \emph{NeurIPS}, 2011.

\bibitem[Plummer et~al.(2016)Plummer, Wang, Cervantes, Caicedo, Hockenmaier,
  and Lazebnik]{plummer2016flickr30k}
Plummer, B.~A., Wang, L., Cervantes, C.~M., Caicedo, J.~C., Hockenmaier, J.,
  and Lazebnik, S.
\newblock Flickr30k entities: Collecting region-to-phrase correspondences for
  richer image-to-sentence models.
\newblock In \emph{arXiv 1505.04870}, 2016.

\bibitem[Qi et~al.(2020)Qi, Su, Song, Cui, Bharti, and Sacheti]{qi:imagebert}
Qi, D., Su, L., Song, J., Cui, E., Bharti, T., and Sacheti, A.
\newblock Imagebert: Cross-modal pre-training with large-scale weak-supervised
  image-text data.
\newblock \emph{arXiv preprintarXiv:2001.07966}, 2020.

\bibitem[Radford et~al.(2021)Radford, Kim, Hallacy, Ramesh, Goh, Agarwal,
  Sastry, Askell, Mishkin, Clark, Krueger, and Sutskever]{radford2021learning}
Radford, A., Kim, J.~W., Hallacy, C., Ramesh, A., Goh, G., Agarwal, S., Sastry,
  G., Askell, A., Mishkin, P., Clark, J., Krueger, G., and Sutskever, I.
\newblock Learning transferable visual models from natural language
  supervision.
\newblock In \emph{arXiv 2103.00020}, 2021.

\bibitem[Rajbhandari et~al.(2019)Rajbhandari, Rasley, Ruwase, and
  He]{DBLP:journals/corr/abs-1910-02054}
Rajbhandari, S., Rasley, J., Ruwase, O., and He, Y.
\newblock Zero: Memory optimization towards training {A} trillion parameter
  models.
\newblock \emph{CoRR}, 2019.

\bibitem[Ramesh et~al.(2021)Ramesh, Pavlov, Goh, Gray, Voss, Radford, Chen, and
  Sutskever]{ramesh2021zeroshot}
Ramesh, A., Pavlov, M., Goh, G., Gray, S., Voss, C., Radford, A., Chen, M., and
  Sutskever, I.
\newblock Zero-shot text-to-image generation.
\newblock In \emph{arXiv 2102.12092}, 2021.

\bibitem[Ryoo et~al.(2021)Ryoo, Piergiovanni, Arnab, Dehghani, and
  Angelova]{ryoo2021tokenlearner}
Ryoo, M.~S., Piergiovanni, A., Arnab, A., Dehghani, M., and Angelova, A.
\newblock Tokenlearner: What can 8 learned tokens do for images and videos?
\newblock In \emph{arXiv 2106.11297}, 2021.

\bibitem[Shao et~al.(2019)Shao, Li, Zhang, Peng, Yu, Zhang, Li, and
  Sun]{Shao_2019_ICCV}
Shao, S., Li, Z., Zhang, T., Peng, C., Yu, G., Zhang, X., Li, J., and Sun, J.
\newblock Objects365: A large-scale, high-quality dataset for object detection.
\newblock In \emph{Proceedings of the IEEE/CVF International Conference on
  Computer Vision (ICCV)}, October 2019.

\bibitem[Sharma et~al.(2018)Sharma, Ding, Goodman, and
  Soricut]{sharma2018conceptual}
Sharma, P., Ding, N., Goodman, S., and Soricut, R.
\newblock Conceptual captions: A cleaned, hypernymed, image alt-text dataset
  for automatic image captioning.
\newblock In \emph{ACL}, pp.\  2556--2565, 2018.

\bibitem[Shen et~al.(2021)Shen, Li, Tan, Bansal, Rohrbach, Chang, Yao, and
  Keutzer]{shen2021much}
Shen, S., Li, L.~H., Tan, H., Bansal, M., Rohrbach, A., Chang, K.-W., Yao, Z.,
  and Keutzer, K.
\newblock How much can clip benefit vision-and-language tasks?
\newblock \emph{arXiv preprint}, 2021.

\bibitem[Tschandl et~al.(2018)Tschandl, Rosendahl, and Kittler]{ham10000}
Tschandl, P., Rosendahl, C., and Kittler, H.
\newblock The ham10000 dataset, a large collection of multi-source
  dermatoscopic images of common pigmented skin lesions.
\newblock \emph{Nature Scientific Data}, 5, 2018.

\bibitem[Vaswani et~al.(2017)Vaswani, Shazeer, Parmar, Uszkoreit, Jones, Gomez,
  Kaiser, and Polosukhin]{NIPS2017_3f5ee243}
Vaswani, A., Shazeer, N., Parmar, N., Uszkoreit, J., Jones, L., Gomez, A.~N.,
  Kaiser, L.~u., and Polosukhin, I.
\newblock Attention is all you need.
\newblock In \emph{Advances in Neural Information Processing Systems},
  volume~30. Curran Associates, Inc., 2017.

\bibitem[Wang et~al.(2020)Wang, Hu, Zhang, Li, Wang, Zhang, Gao, and
  Liu]{abs-2012-06946}
Wang, J., Hu, X., Zhang, P., Li, X., Wang, L., Zhang, L., Gao, J., and Liu, Z.
\newblock Minivlm: {A} smaller and faster vision-language model.
\newblock \emph{arXiv preprint arXiv:2012.06946}, 2020.

\bibitem[Wang et~al.(2017)Wang, Peng, Lu, Lu, Bagheri, and Summers]{chestx}
Wang, X., Peng, Y., Lu, L., Lu, Z., Bagheri, M., and Summers, R.~M.
\newblock Chestx-ray8: Hospital-scale chest x-ray database and benchmarks on
  weakly-supervised classification and localization of common thorax diseases.
\newblock In \emph{arXiv 1705.02315}, 2017.

\bibitem[Wang et~al.(2021)Wang, Yu, Yu, Dai, Tsvetkov, and Cao]{wang2021simvlm}
Wang, Z., Yu, J., Yu, A.~W., Dai, Z., Tsvetkov, Y., and Cao, Y.
\newblock Simvlm: Simple visual language model pretraining with weak
  supervision.
\newblock In \emph{arXiv 2108.10904}, 2021.

\bibitem[Wu et~al.(2021)Wu, Xiao, Codella, Liu, Dai, Yuan, and
  Zhang]{Wu_2021_ICCV}
Wu, H., Xiao, B., Codella, N., Liu, M., Dai, X., Yuan, L., and Zhang, L.
\newblock Cvt: Introducing convolutions to vision transformers.
\newblock In \emph{Proceedings of the IEEE/CVF International Conference on
  Computer Vision (ICCV)}, pp.\  22--31, October 2021.

\bibitem[Xie et~al.(2020)Xie, Luong, Hovy, and Le]{Xie_2020_CVPR}
Xie, Q., Luong, M.-T., Hovy, E., and Le, Q.~V.
\newblock Self-training with noisy student improves imagenet classification.
\newblock In \emph{Proceedings of the IEEE/CVF Conference on Computer Vision
  and Pattern Recognition (CVPR)}, June 2020.

\bibitem[Xu et~al.(2021{\natexlab{a}})Xu, Ghosh, Huang, Okhonko, Aghajanyan,
  Metze, Zettlemoyer, and Feichtenhofer]{xu2021videoclip}
Xu, H., Ghosh, G., Huang, P.-Y., Okhonko, D., Aghajanyan, A., Metze, F.,
  Zettlemoyer, L., and Feichtenhofer, C.
\newblock Videoclip: Contrastive pre-training for zero-shot video-text
  understanding.
\newblock In \emph{EMNLP}, 2021{\natexlab{a}}.

\bibitem[Xu et~al.(2016)Xu, Mei, Yao, and Rui]{xu2016msr}
Xu, J., Mei, T., Yao, T., and Rui, Y.
\newblock Msr-vtt: A large video description dataset for bridging video and
  language.
\newblock In \emph{CVPR}, pp.\  5288--5296, 2016.

\bibitem[Xu et~al.(2021{\natexlab{b}})Xu, Zhang, Hu, Wang, Wang, Wei, Bai, and
  Liu]{Xu_2021_ICCV}
Xu, M., Zhang, Z., Hu, H., Wang, J., Wang, L., Wei, F., Bai, X., and Liu, Z.
\newblock End-to-end semi-supervised object detection with soft teacher.
\newblock In \emph{Proceedings of the IEEE/CVF International Conference on
  Computer Vision (ICCV)}, pp.\  3060--3069, October 2021{\natexlab{b}}.

\bibitem[Xue et~al.(2021)Xue, Huang, Liu, Peng, Fu, Li, and
  Luo]{xue2021probing}
Xue, H., Huang, Y., Liu, B., Peng, H., Fu, J., Li, H., and Luo, J.
\newblock Probing inter-modality: Visual parsing with self-attention for
  vision-language pre-training.
\newblock In \emph{NeurIPS}, 2021.

\bibitem[Yang et~al.(2021)Yang, Li, Zhang, Dai, Xiao, Yuan, and
  Gao]{yang2021focal}
Yang, J., Li, C., Zhang, P., Dai, X., Xiao, B., Yuan, L., and Gao, J.
\newblock Focal self-attention for local-global interactions in vision
  transformers.
\newblock In \emph{arXiv 2107.00641}, 2021.

\bibitem[Yang et~al.(2022)Yang, Li, Zhang, Xiao, Liu, Yuan, and
  Gao]{Jianwei_UNICL2022}
Yang, J., Li, C., Zhang, P., Xiao, B., Liu, C., Yuan, L., and Gao, J.
\newblock Unified contrastive learning in image-text-label space.
\newblock In \emph{arXiv In Preparation}, 2022.

\bibitem[Yao et~al.(2021)Yao, Huang, Hou, Lu, Niu, Xu, Liang, Li, Jiang, and
  Xu]{yao2021filip}
Yao, L., Huang, R., Hou, L., Lu, G., Niu, M., Xu, H., Liang, X., Li, Z., Jiang,
  X., and Xu, C.
\newblock Filip: Fine-grained interactive language-image pre-training.
\newblock In \emph{arXiv 2111.07783}, 2021.

\bibitem[Yu et~al.(2020)Yu, Tang, Yin, Sun, Tian, Wu, and Wang]{yu:ernie-vil}
Yu, F., Tang, J., Yin, W., Sun, Y., Tian, H., Wu, H., and Wang, H.
\newblock Ernie-vil: Knowledge enhanced vision-language representations through
  scene graph.
\newblock \emph{arXiv preprint arXiv:2006.16934}, 2020.

\bibitem[Yu et~al.(2018)Yu, Kim, and Kim]{yu2018joint}
Yu, Y., Kim, J., and Kim, G.
\newblock A joint sequence fusion model for video question answering and
  retrieval.
\newblock In \emph{ECCV}, pp.\  471--487, 2018.

\bibitem[Zhai et~al.(2021)Zhai, Kolesnikov, Houlsby, and
  Beyer]{zhai2021scaling}
Zhai, X., Kolesnikov, A., Houlsby, N., and Beyer, L.
\newblock Scaling vision transformers.
\newblock In \emph{arXiv 2106.04560}, 2021.

\bibitem[Zhang et~al.(2021{\natexlab{a}})Zhang, Dai, Yang, Xiao, Yuan, Zhang,
  and Gao]{zhang2021multi}
Zhang, P., Dai, X., Yang, J., Xiao, B., Yuan, L., Zhang, L., and Gao, J.
\newblock Multi-scale vision longformer: A new vision transformer for
  high-resolution image encoding.
\newblock \emph{ICCV 2021}, 2021{\natexlab{a}}.

\bibitem[Zhang et~al.(2021{\natexlab{b}})Zhang, Li, Hu, Yang, Zhang, Wang,
  Choi, and Gao]{Zhang_2021_CVPR}
Zhang, P., Li, X., Hu, X., Yang, J., Zhang, L., Wang, L., Choi, Y., and Gao, J.
\newblock Vinvl: Revisiting visual representations in vision-language models.
\newblock In \emph{Proceedings of the IEEE/CVF Conference on Computer Vision
  and Pattern Recognition (CVPR)}, pp.\  5579--5588, June 2021{\natexlab{b}}.

\bibitem[Zhou et~al.(2021)Zhou, Koltun, and Krähenbühl]{zhou2021simple}
Zhou, X., Koltun, V., and Krähenbühl, P.
\newblock Simple multi-dataset detection.
\newblock In \emph{arXiv 2102.13086}, 2021.

\bibitem[Zoph et~al.(2020)Zoph, Ghiasi, Lin, Cui, Liu, Cubuk, and
  Le]{ZophGLCLC020_nips}
Zoph, B., Ghiasi, G., Lin, T.-Y., Cui, Y., Liu, H., Cubuk, E.~D., and Le, Q.
\newblock Rethinking pre-training and self-training.
\newblock In \emph{NeurIPS}, 2020.

\end{thebibliography}
\bibliographystyle{icml2021}

\end{document}